\documentclass[a4paper,fleqn]{cas-dc}

\usepackage[authoryear]{natbib}

\def\tsc#1{\csdef{#1}{\textsc{\lowercase{#1}}\xspace}}
\tsc{WGM}
\tsc{QE}

\begin{document}
\let\WriteBookmarks\relax
\def\floatpagepagefraction{1}
\def\textpagefraction{.001}

\shorttitle{Scalar Field based Toolpath planning}    

\shortauthors{CQ Shen et~al.}

\clearpage 
\thispagestyle{empty} 
\clearpage 
\setcounter{page}{1} 

\title [mode = title]{Topology-Preserving Scalar Field Optimization for Boundary-Conforming Spiral Toolpaths on Multiply Connected Freeform Surfaces}



%

\author[1]{Changqing Shen}
\ead{D202280361@hust.edu.cn}
\credit{Conceptualization, Methodology, Software, Validation, Formal analysis, Investigation, Visualization, Writing - Original Draft}

\author[1]{Bingzhou Xu}
\ead{joyeuxxu@hust.edu.cn}
\credit{Investigation, Validation, Writing - Review \& Editing}

\author[1]{Bosong Qi}
\ead{poisson_bs@hust.edu.cn}
\credit{Investigation, Validation, Writing - Review \& Editing}

\author[1]{Xiaojian Zhang}
\ead{xjzhang@hust.edu.cn}
\cormark[1]
\cortext[cor1]{Corresponding author}
\credit{Conceptualization, Supervision, Project administration, Funding acquisition, Resources, Writing - Review \& Editing}

\author[1]{Han Ding}
\ead{dinghan@hust.edu.cn}
\credit{Supervision, Project administration, Funding acquisition, Resources}

\affiliation[1]{organization={State Key Laboratory of Intelligent Manufacturing Equipment and Technology, School of Mechanical Science and Engineering},
            addressline={Huazhong University of Science and Technology}, 
            city={Wuhan},
            postcode={430074}, 
            state={Wuhan},
            country={China}}

\begin{abstract}
  Multiply connected freeform surface features are widely encountered in industrial components, where toolpath generation often suffers from discontinuities, sharp turns, non-uniform scallop heights, and incomplete boundary coverage. This paper proposes a scalar-field variational optimization method for milling that produces continuous, boundary-conforming, and non-self-intersecting toolpaths with smoother transitions, more uniform spacing, and reduced redundant path length. A feasible singularity-free initial scalar field with boundary-conforming iso-level sets is first constructed via conformal slit mapping. The optimization is then reformulated as a topology-preserving mesh deformation process governed by boundary-synchronous updates, whereby the continuity, boundary-conformity, and non-self-intersection requirements of the toolpath are converted into mesh-shape constraints maintained throughout the iterative optimization. As a result, the proposed method achieves globally optimized path spacing and improved scallop-height uniformity while preserving trajectory smoothness. Milling experiments show that, compared with a state-of-the-art conformal slit mapping-based method, the proposed approach improves machining efficiency by 14.24\%, enhances scallop-height uniformity by 5.70\%, and reduces milling impact-induced vibrations by over 10\%. The proposed strategy provides an effective solution for high-performance machining of complex multiply connected freeform components.
  
\end{abstract}


\maketitle

\section{Introduction}
\subsection{Problem Description}
\begin{figure}
  \centering
  \includegraphics[width=0.8\linewidth]{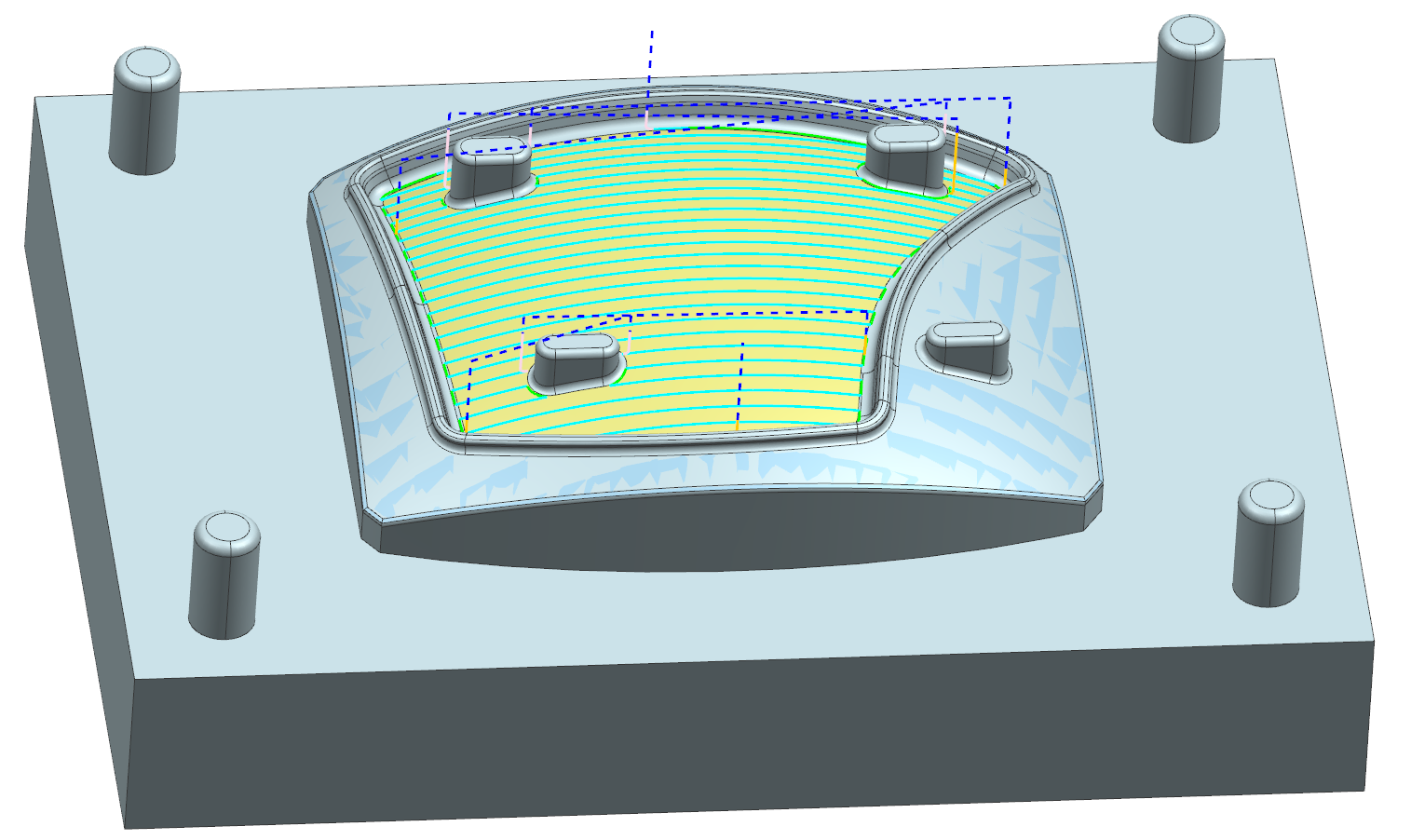}
  \caption{Toolpaths generated for freeform surface machining using commercial CAD/CAM software.}
  \label{fig:headlamp}
\end{figure}

In modern manufacturing, component geometries have become increasingly complex, particularly in the automotive and aerospace sectors\cite{Mali2021A,Marin2025Dimensional,Pereira2025EnhancingPO}. Machining quality and efficiency depend strongly on the quality of toolpath planning. A representative example is an automotive headlamp mold (Figure~\ref{fig:headlamp}), whose machined surface contains multiple protrusions. These geometric features obstruct continuous toolpath coverage, causing conventional planning methods to generate five retractions and numerous sharp turns. These path characteristics tend to induce surface defects and reduce machining efficiency. Studies have examined the adverse machining consequences induced by unfavorable toolpath characteristics. For instance, intermittent tool–workpiece separation and re-engagement can lead to fluctuations in material removal \cite{Yan2019An,Yang2024Dynamic} and burr formation \cite{Chen2021Investigation}; frequent steering operations may trigger actuator deceleration \cite{Kim2002Machining} and intensified vibrations \cite{Wang2025Tool}; toolpaths that do not conform to surface boundaries tend to leave residual material \cite{Yang2023Template}; and self-intersecting trajectories can distort the milling load and cause surface scratching on the machined workpiece \cite{Hajdu2025Stable}.

To satisfy the requirements of advanced complex components for both high efficiency and high quality, toolpath planning should minimize path length, interruptions, sharp turns, and self-intersections while enhancing smoothness, scallop-height uniformity, and strict conformity to machining surface boundaries. However, these objectives are often mutually conflicting, which renders their simultaneous achievement within a unified global framework particularly challenging.

\subsection{Related work}
Based on current knowledge, milling toolpath planning methods for multiply connected surfaces can be broadly classified into three categories: element decomposition \cite{Acar2003Path,Choi2021Efficient,Ge2024Spiral,Held2018On,Lin2017Robot,Sun2016Smooth,Wu2019Energy,Xu2022Global}, boundary offsetting \cite{Lee2003Contour,Xu2019Contour,Zhao2024Hybrid,Zhuang2010High}, and scalar/vector field optimization \cite{Bartoň2021Geometry,Dutta2023Vector,Goes2016Vector,Shen2024Spiral,Shen2025Conformal,Zou2021Length,Zou2014Iso}. In addition, several complete coverage trajectory planning methods have been proposed but are rarely adopted in milling applications because of excessive turning angles, incomplete coverage, and frequent self-intersections \cite{Ban2013Topology,Song2018Epsilon}; therefore, they are not discussed in this paper. The first category decomposes complex surfaces into multiple simpler subregions that can be more easily parameterized and are typically assumed to be connected. For example, Acar et al. \cite{Acar2003Path} divided complex regions by identifying critical boundary points and sweep lines, allowing each subregion to be fully covered by sweep paths. Yu-Yao Lin et al. \cite{Lin2017Robot} employed holomorphic quadratic differentials to decompose a genus-g surface into 3g-3 hole-free subregions, each filled with zigzag coverage paths. Other studies have introduced artificial cutting boundaries to transform multiply connected domains into single or multiple disk- or annulus-like subregions \cite{Sun2016Smooth,abrahamsen2019cuthole}, within which conformal mapping \cite{Choi2021Efficient} or medial-axis-transform (MAT)-based methods \cite{Ge2024Spiral,Held2018On} were adopted to generate spiral or zigzag paths. However, for highly perforated surfaces, such decomposition is often nontrivial and can lead to excessive bridging or severe non-uniformity in path spacing.

The second category is based on boundary offsetting. Lee et al. \cite{Lee2003Contour} proposed an equal-scallop-height offsetting method to generate spiral paths on hole-free surfaces. Xu et al. \cite{Xu2019Contour} computed equal-scallop-height offsets on flattened surfaces using mapping stretch factors, achieving the improved smoothness compared with Lee’s method. Nevertheless, both approaches suffer from sharp turning angles and fragmented paths caused by self-intersections induced by trajectory offsetting. Zhao et al. \cite{Zhao2024Hybrid} extracted sharp-turning regions from offset paths and filled them with zigzag patterns to improve the spacing uniformity, at the cost of increased turns and interruptions. Zhuang et al. \cite{Zhuang2010High} formulated the offsetting process using the level-set evolution, which enabled the smoothness control, while on multiply connected surfaces, the level sets tend to fragment, necessitating additional path bridging.

Regardless of whether decomposition- or offsetting-based strategies are adopted, existing methods frequently generate toolpaths with interruptions, sharp turning angles, and insufficient boundary conformity when holes and protrusions are present. These limitations inevitably degrade machining quality and efficiency, highlighting the need for new toolpath planning approaches capable of producing continuous, smooth, and fully boundary-conforming trajectories on complex multiply connected surfaces.

Recently, scalar/vector field optimization methods \cite{Bartoň2021Geometry,Dutta2023Vector,Goes2016Vector,Shen2024Spiral,Shen2025Conformal,Zou2021Length,Zou2014Iso} have emerged as promising alternatives because they provide a global optimization framework that simultaneously accounts for path spacing uniformity, smoothness, and boundary conformity.

\begin{figure}
  \centering
  \includegraphics[width=0.8\linewidth]{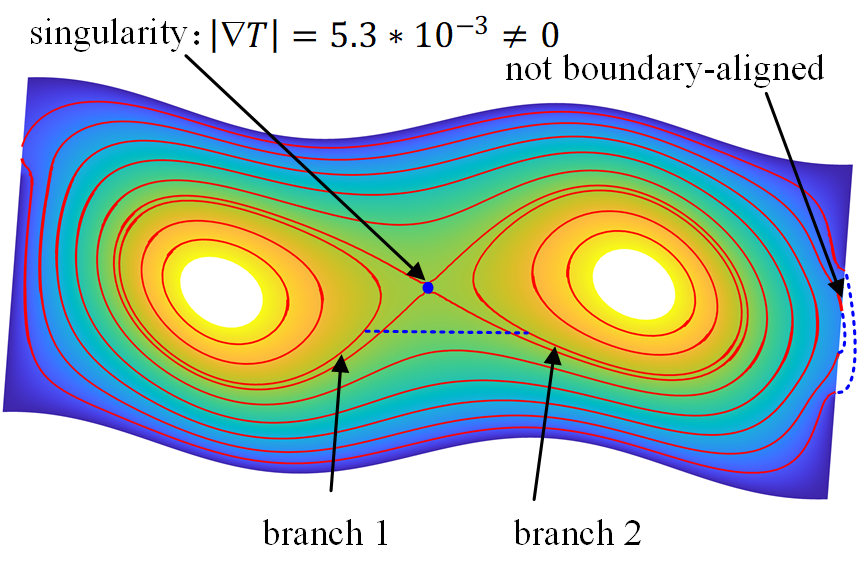}
  \caption{Branching and discontinuities of toolpaths on multiply connected surfaces when the required scalar-field constraints are not satisfied.}
  \label{fig:singularity}
\end{figure}

Zou et al. \cite{Zou2014Iso} and Zou \cite{Zou2021Length} formulated toolpath planning on hole-free surfaces as a scalar-field functional optimization problem, in which multi-objective energy terms were used to optimize isoparametric trajectories. Their work showed that, with appropriate boundary conditions and nonvanishing-gradient constraints, boundary-aligned and globally coordinated trajectories could be generated from scalar-field iso-level sets. However, extending this framework to porous or multiply connected surfaces remains challenging, because simultaneously maintaining boundary conformity and a nonzero gradient sigularity \cite{Mitra2024Singular,Campen2016Bijective,Vekhter2019Weaving,Zhang2006Vector} throughout numerical optimization is difficult \cite{Li2006Representing,Makhanov2022Vector,Wu2024Pose}. As a result, the optimization of scalar fields on such surfaces still requires further investigation. Figure~\ref{fig:singularity} illustrates the branching and discontinuities that may arise when these conditions are not satisfied.

Building on this concept, Shen et al. \cite{Shen2024Spiral,Shen2025Conformal} employed conformal slit mapping \cite{Li2025A,Nasser2011Numerical,Wu2023Numerical} to construct scalar fields with nonzero gradients and fixed boundary isovalues on multiply connected surfaces. This mapping transformed a complex, multiply connected surface into a simply connected disk or annulus, in which internal holes were mapped to circular arcs, thereby facilitating the construction of scalar fields that satisfied the prescribed dual constraints. Under these constraints, spiral trajectories derived from the scalar field remained continuous, boundary-conforming, and smoothly bridged, whereas the rigid nature of conformal slit mapping can inherently restrict the freedom of subsequent optimization.

Some other studies that do not require the surface to contain holes and the toolpath to conform to the boundary include the following. Hao et al. \cite{Hao2025Neural} introduced a physics-informed neural network (PINN) to optimally partition surfaces and generate toolpaths guided by a multi-objective scalar field loss function. Chen et al. \cite{Chen2025Singularity} further incorporated actuator dynamic characteristics into the scalar-field loss formulation. In addition, M. Bartoň et al. \cite{Bartoň2021Geometry} evaluated the potential interference between non-spherical tools and the workpiece surface using a wedge diagram and incorporated an interference-avoidance term into the scalar-field energy function. The resulting optimal scalar field, solved via TensorFlow, enabled multi-objective optimization that simultaneously avoided tool interference, ensured scallop-height uniformity, and improved toolpath fairness.

These neural network-driven techniques have shown promise in addressing the nonlinear functional optimization of complex scalar fields. However, to the best of our knowledge, they still suffer from poor convergence under complex boundary conditions and incur substantial computational costs \cite{Grossmann2024Can}. For instance, for surfaces with first-order discontinuity boundaries or internal holes, strict boundary conformity requires scalar fields with derivative-discontinuous behavior near the boundaries. Recent studies \cite{Della2023Discontinuous,Rahaman2019On} have suggested that neural networks exhibit a low-frequency bias towards smooth approximations, which limits their accuracy in representing such discontinuous functions.

In addition, early work by E. Bohez et al. \cite{Bohez2000Adaptive} proposed a functional optimization objective emphasizing global scallop-height uniformity and toolpath smoothness. However, it did not establish an explicit relationship between multiple optimization objectives and the local differential geometric properties of the scalar field on the surface. Consequently, the method was restricted to quadrilateral-meshed Coons surfaces and did not support adaptive path refinement.

In summary, scalar-field optimization methods for toolpath planning enable multi-objective consideration within a global optimization framework. Nevertheless, effectively optimizing scalar fields on multiply connected surfaces under nonzero-gradient constraints while maintaining strict boundary conformity to further improve path quality remains a significant challenge.

\subsection{Our Contribution}
In this study, a scalar-field-based functional optimization method was developed for multiply connected surfaces to generate globally optimized spiral toolpaths. The resulting paths exhibit uninterrupted continuity, strict boundary conformity, smooth turning transitions, and uniform scallop height. These capabilities are achieved through the following two technical contributions:
\begin{enumerate}
    \item A conformal slit mapping is employed to construct an initial scalar field that satisfies the prescribed dual constraints of boundary conformity and nonzero gradients while remaining close to the optimal solution. This initialization ensures that the optimization starts from a constraint-compliant, near-optimal state and substantially reduces the computational cost by avoiding invalid iterations.

    \item The scalar-field iteration is reformulated from updating vertex scalar values to optimizing the mesh geometry. Mesh geometry constraints are imposed to strictly eliminate internal singularities and ensure uniform boundary values of the scalar field, thereby preserving the boundary-conforming, continuity, and non-self-intersection of the scalar field’s isoparametric lines during iterative optimization.
\end{enumerate}

\section{Method}
\subsection{Evaluation Criteria for Scalar-Field Level-Set Paths}

\begin{figure}
  \centering
  \includegraphics[width=0.8\linewidth]{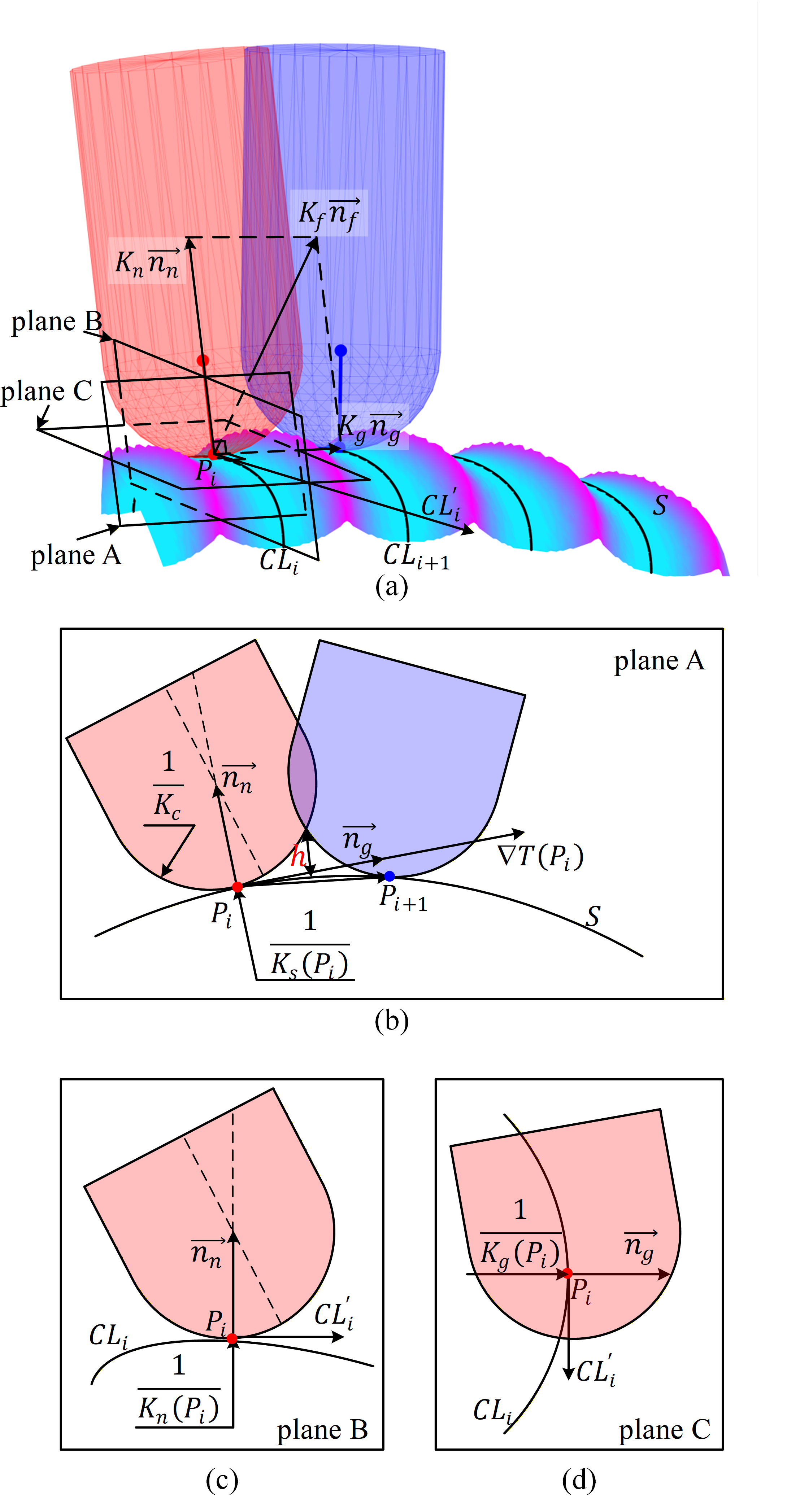}
  \caption{\label{fig:relationships}Relationship among $CL_i$, $CL_{i+1}$, $h$, $\nabla T(P_i)$, $K_s$, $K_c$, $K_f$, $K_n$, and $K_g$.}
\end{figure}

One of the key criteria for evaluating a coverage trajectory is the uniformity of the spacing between neighboring paths. In general coverage problems, this is often measured by the Euclidean distance between adjacent trajectories, where more uniform spacing usually implies less redundant coverage. In ball-end milling, a closely related but more meaningful quantity is the scallop height $h$ left between neighboring toolpaths, as shown in Figure~\ref{fig:relationships}. Therefore, instead of enforcing uniform Euclidean spacing directly, the objective is to maintain a uniform scallop height over the surface.

When the toolpaths are given by the iso-level curves of a scalar field $T$, the scallop height between two adjacent iso-level curves can be approximated by
\begin{equation}
h = |T_{i+1}-T_i|^2 \frac{K_s+K_c}{8\,\|\nabla T(P_i)\|^2_2}
\label{eq:scallop_height_simple}
\end{equation}

where $P_i$ is a point on the surface $S$, and $|T_{i+1}-T_i|$ denotes the absolute scalar difference between two adjacent iso-level curves. During toolpath generation, this quantity can be treated as a constant. $K_c$ is the curvature of the ball-end cutter, and $K_s$ denotes the normal curvature of $S$ at $P_i$ along the gradient direction $\nabla T$. Based on this observation, the global uniformity of scallop height can be evaluated by integrating a symmetrized Itakura--Saito-type divergence \cite{Itakura2020} over the surface:
\begin{equation}
E_w=\frac{1}{|S|}\int_S \left(
\frac{K_s+K_c}{8\,Q_S\,\|\nabla T\|_2^2}
+\frac{8\,Q_S\,\|\nabla T\|_2^2}{K_s+K_c}
\right)\, dS
\label{eq:Ew}
\end{equation}

where $|S|$ denotes the area of the surface $S$, and $Q_S$ denotes the mean value of a metric proportional to the scallop height over $S$:
\begin{equation}
Q_S=\frac{1}{|S|}\int_S \left(\frac{K_s+K_c}{8\,\|\nabla T\|_2^2}\right)\, dS
\label{eq:Qs}
\end{equation}

A detailed derivation of Equation~\ref{eq:scallop_height_simple} and the discrete form of $Q_S$ and Equation~\ref{eq:Ew} on a triangular mesh are provided in the Appendix and in \cite{Zou2014Iso}.
It is worth noting that global scallop-height uniformity could also be measured by the sum of square between the actual scallop height and a prescribed target value \cite{Zou2014Iso}. However, the advantage of Equation~\ref{eq:Ew} is twofold. First, it is invariant to a global scaling of $T$. Second, it imposes a stronger penalty as $\|\nabla T\|$ approaches zero, which helps accelerate convergence in the subsequent optimization and prevents excessively dense toolpaths. In practical simulations, setting $Q_S$ was found to still provide satisfactory scallop-height equalization. In this case, Equation~\ref{eq:Qs} drives the scallop-height-related metric toward the prescribed constant value rather than its surface average, thereby avoiding the surface integral used to evaluate $Q_S$ and reducing the computational time by approximately 20\%.

Subsequently, another key criterion is the smoothness of the iso-level toolpaths. It can be quantified by the surface integral of its squared curvature:
\begin{equation}
    E_k=\int_SK_f^2dS
    \label{eq:Ek}
\end{equation}

where $K_f$ denotes the curvature of the iso-level toolpath at a point on $S$. It can be decomposed into the normal curvature $K_n$ and the geodesic curvature $K_g$. The discrete form of Equation ~\ref{eq:Ek} on a triangular mesh is provided in the Appendix.

A comprehensive metric $E$ is defined to jointly evaluate milling scallop-height uniformity and toolpath smoothness as follows:

\begin{equation}
    E=E_w+\alpha E_k
    \label{eq:Eall}
\end{equation}

where $\alpha$ is a weighting parameter that balances scallop-height uniformity and toolpath smoothness, which is selected according to the specific machining conditions.

\subsection{Scalar Field Initialization Based on Conformal Slit Mapping}
\begin{figure}
  \centering
  \includegraphics[width=0.8\linewidth]{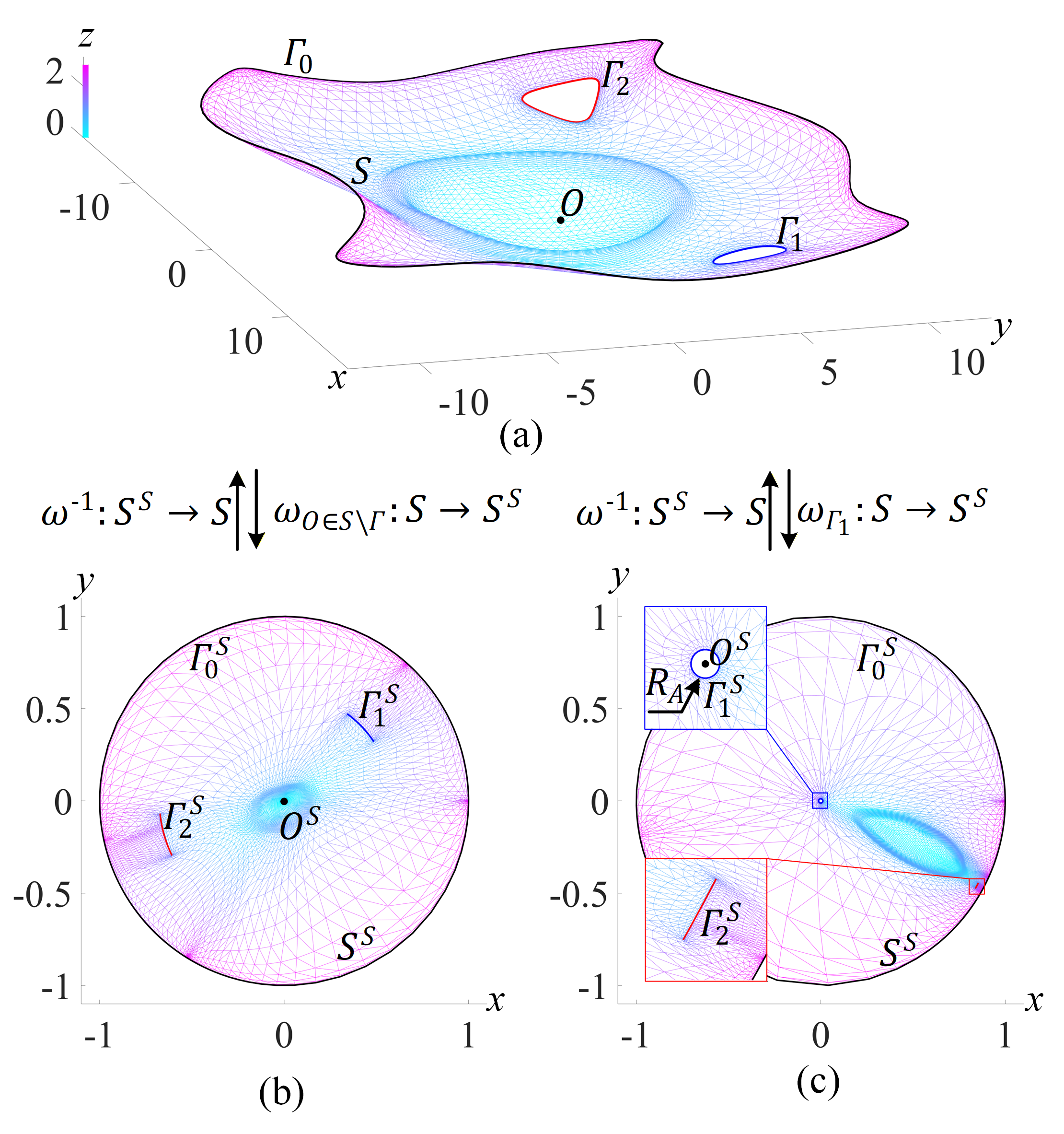}
  \caption{\label{fig:TwoCSM}Two types of conformal slit mapping. (a) Original surface, (b) disk conformal slit mapping result, and (c) annular conformal slit mapping result.}
\end{figure}

As shown in Figure~\ref{fig:TwoCSM}(a), let the boundary components of a multiply connected surface $S$ be denoted by $\Gamma=\Gamma_0+\Gamma_1+\Gamma_2+\ldots+\Gamma_m$, where $m>1$. Conformal slit mapping is used to construct an initial scalar field $T_{\mathrm{init}}$ that approximates the optimal solution of $T$. The field $T_{\mathrm{init}}$ satisfies $\|\nabla T_{\mathrm{init}}\|>0$ on the open domain $S \setminus\Gamma$ and $T_{\mathrm{init}}(\Gamma_i)=C_i$ for $i=0,1,\ldots,m$, where $C_i$ are undetermined constants. Consequently, the level sets of $T_{\mathrm{init}}$ conform to the surface boundaries and remain free of discontinuities and self-intersections, providing a feasible initialization for the subsequent optimization.

Specifically, if an interior point $O\in S\setminus\Gamma$ is selected together with the boundary $\Gamma_0$, a disk conformal slit mapping transforms $S$ into a unit disk domain $D^S$ with circular-arc slits, as shown in Figure~\ref{fig:TwoCSM}(b). Under this mapping, $\Gamma_0$ is mapped to the outer boundary of the unit disk, $O$ is mapped to the disk center $O^S$, and each remaining boundary $\Gamma_i$ $(i=1,\ldots,m)$ is mapped to an arc-shaped slit. Alternatively, if two boundary components $\Gamma_0$ and $\Gamma_1$ are selected, an annular conformal slit mapping transforms $S$ into a unit annular domain $A^S$ with circular-arc slits, as shown in Figure~\ref{fig:TwoCSM}(c). In this case, $\Gamma_0$ and $\Gamma_1$ are mapped to the outer and inner boundaries of the annulus, respectively, while the remaining boundaries $\Gamma_i$ $(i=2,\ldots,m)$ are mapped to arc-shaped slits.

In this work, the considered machining surfaces are open surface patches with multiple boundary components, where one boundary is typically in contact with the worktable or connected to the rest of the workpiece. Therefore, $\Gamma_0$ was designated as this boundary. For convenience, both disk and annular mappings were uniformly denoted as $\omega$, and the corresponding mapped domain was denoted as $S^S$, expressed as:

\begin{equation}
\begin{cases} 
\omega_{O \in S \setminus \Gamma}: S \to S^S & \text{for disk conformal map} \\
\omega_{\Gamma_i,\,i=1,2,\dots,m}: S \to S^S & \text{for annular conformal map}
\end{cases}
\end{equation}

The scalar field $T^S$ on $S^S$ is defined as follows:

\begin{equation}
    T^S=f(\| S^S \|)
\end{equation}

where $\|S^S\|$ denotes the distance from a point in $S^S$ to $O^S$. In particular, this distance is constant along each boundary component. The function $f$ is required to be monotonically increasing on $[\min(\|S^S\|),1]$:

\begin{equation}
f: \left[ \min\left( \|S^S\| \right), 1 \right] \to \mathbb{R}, f'(x) > 0 \ 
\label{eq:fconstraint}
\end{equation}

where $\ x \in \left[ \min\left( \|S^S\| \right), 1 \right]$. For disk conformal mapping, $\min(\|S^S\|)=0$, whereas for annular conformal mapping, $\min(\|S^S\|)=R_A$, where $R_A$ denotes the radius of the inner boundary of the annulus.

Because $S^S$ and $S$ are topologically equivalent, the nodal scalar values of $T^S$ can be assigned to the corresponding nodes of $S$, thereby yielding the scalar field $T$ on $S$. Hence, $T$ is determined by the unknown monotonically increasing function $f$ and an element $\Theta$, where $\Theta \in \{S \setminus \Gamma\} \cup \{\Gamma_i \mid i=1,2,\dots,m\}$. Specifically, $\Theta$ can be either an interior point of $S$, in which case a disk conformal slit mapping is determined, or a boundary component of $S$, in which case an annular conformal slit mapping is determined. Once $\Theta$ is given, the corresponding conformal slit mapping is determined. For a given pair $(\Theta,f)$, the scalar field is written as $T(\Theta,f):S\to\mathbb{R}$, and the corresponding energy is denoted by $E(T(\Theta,f))$.

The energy $E$ is jointly determined by $\Theta$ and $f$. The monotonicity of $f$ guarantees the existence of a nonzero radial gradient component on $S^S$, thereby preventing the constructed scalar field from containing zero-gradient singularities.

The objective is to minimize $E$ with respect to both $\Theta$ and $f$; however, direct joint optimization is difficult. Therefore, $\Theta$ is first fixed, and the optimal $f$ is determined by a perturbation-based method, yielding the corresponding energy value. The optimal location of $\Theta$ is then searched over the surface along the gradient direction of $E$. Once $\Theta$ reaches its optimal position $\Theta^{\mathrm{opt}}$, the corresponding $f_{\min}$ is obtained, and $T\left(\Theta^{\mathrm{opt}}, f_{\min}\right)$ is taken as the desired initial scalar field $T_{\mathrm{init}}$, for which $E(T_{\mathrm{init}})=E_{\min}^{\mathrm{opt}}$. More details can be found in the Appendix and in \cite{Shen2025Conformal}.

\subsection{Variational Optimization Method for Scalar-Field Level-Set Toolpaths}

As discussed in Section 2.2, conformal slit mapping provides a feasible initial scalar field $T_{\mathrm{init}}$ with a relatively low energy $E$. However, the iso-level curves of $T_{\mathrm{init}}$ are restricted to those generated from concentric circles in $S^S$, and this restriction is introduced only for initialization rather than being an essential constraint of the final optimization problem. Therefore, further optimization is still possible in a broader function space. Starting from $T_{\mathrm{init}}$, the final optimization is formulated over the admissible set
\begin{equation}
\begin{cases}
E_{\text{opt}} := \min_{\substack{T \in \{\|\nabla T\| > 0 \ \text{on}\ S \setminus \Gamma, \\ T(\Gamma_i) = C_i \ \text{for}\ i=0,1,\dots,m\}}} E \\
T_{\text{opt}} := \arg\min_{\substack{T \in \{\|\nabla T\| > 0 \ \text{on}\ S \setminus \Gamma, \\ T(\Gamma_i) = C_i \ \text{for}\ i=0,1,\dots,m\}}} E
\end{cases}
\label{eq:itertarget}
\end{equation}

The procedure is as follows. Let the initial value of $T$ be $T_{init}$. A suitable perturbation $\delta T$ is then determined such that $E(T+\delta T)< E(T)$, while ensuring that $T+\delta T$ continues to satisfy the imposed constraints:

\begin{equation}
\begin{aligned}
\|\nabla(T+\delta T)\| &> 0
\quad \text{on } S \setminus \Gamma, \\
(T+\delta T)(\Gamma_i) &= C_i + \delta C_i,
\quad i=0,1,\dots,m
\end{aligned}
\label{eq:constraint}
\end{equation}

By repeatedly replacing $T$ with $T+\delta T$, the scalar field $T$ iteratively converges to the optimal solution $T_{opt}$.

\subsubsection{Scalar-Field Iterative Constraints Derived from Mesh Topology}

\begin{figure}
  \centering
  \includegraphics[width=0.8\linewidth]{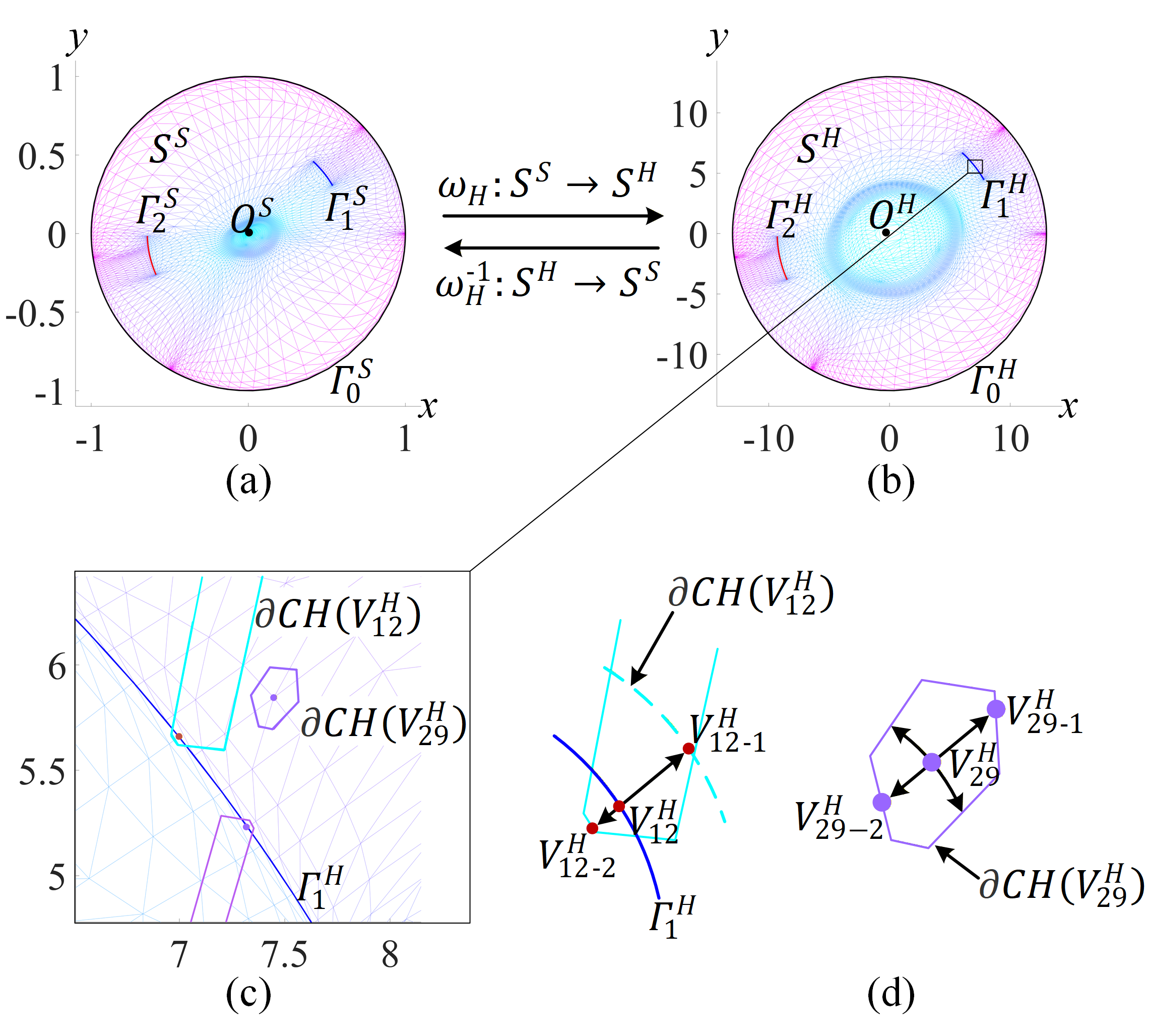}
  \caption{\label{fig:meshInitialization}Mesh initialization and iterative constraints. (a) Conformal slit mapping mesh, (b) initialized mesh, (c) convex hull corresponding to vertices, and (d) iterative constraints of vertices corresponding to the convex hull.}
\end{figure}

Based on the topological equivalence between $S$ and $S^S$, the initial scalar field $T_{\mathrm{init}}$ can be represented on $S^S$ as $T_{\mathrm{init}}^S$. As shown in Figures~\ref{fig:meshInitialization}(a) and (b), the domain $S^S$ is then transformed by a homeomorphism $\omega_H$ \cite{Shen2019Networks} into a new slit-containing disk or annular domain, denoted by $S^H$:
\begin{equation}
\omega_H: S^S \rightarrow S^H:
S_i^H = \frac{S_i^S}{\|S_i^S\|}\, T_{\mathrm{init}}^S(S_i^S), \quad \text{for } S_i^S \in S^S.
\label{eq:initMesh}
\end{equation}
That is, each point $S_i^S$ is radially relocated so that its distance from $O^S$ equals $T_{\mathrm{init}}^S(S_i^S)$, as illustrated in Figures~\ref{fig:meshInitialization}(a) and (b).

Next, the distance from each mesh node to the center $O^H$ is taken as its scalar value. Under this definition, the variation of the nodal scalar values is equivalent to the radial displacement of the nodes. Accordingly, the scalar field $T^H$ on $S^H$ is defined by
\begin{equation}
T^H = \|S^H\|.
\label{eq:TH}
\end{equation}

Because $S^H$ is topologically equivalent to $S$, the scalar field $T^H$ can be mapped back to $S$, thereby yielding the scalar field $T$. In this way, the optimization of $T$ is equivalently transformed into an iterative update of mesh node positions on $S^H$, whose initial configuration is given by Equation~\ref{eq:initMesh}.

As illustrated in Figure~\ref{fig:meshInitialization}(c), $S^H$ is represented as a triangular mesh, where $V^H$ and $F^H$ denote the sets of vertices and faces, respectively. Therefore, the optimization on $S^H$ can be described by the motion of $V^H$. Consequently, the constraints in Equation~\ref{eq:constraint} can be reformulated as two topology-related constraints on the mesh of $S^H$:

\textbf{Constraint 1}. During the iteration on $S^H$, the mesh topology must remain homeomorphic to that of $S$; that is, no triangular face in $S^H$ is allowed to invert. This condition ensures that $\|\nabla T\| > 0$ on $S\setminus\Gamma$.

\textbf{Constraint 2}. Each boundary $\Gamma_i^H$ on $S^H$ must remain a circle or circular arc centered at $O^H$. This condition ensures that
\[
T(\Gamma_i)=T^H(\Gamma_i^H)=\|\Gamma_i^H\|=C_i,\qquad i=0,1,\ldots,m.
\]

As shown in Figure~\ref{fig:meshInitialization}(d), each vertex $V_i^H\in V^H$ has two degrees of freedom, corresponding to the radial and circumferential directions on $S^H$. According to Equation~\ref{eq:TH}, only radial motion of $V_i^H$ affects $T^H$, whereas circumferential motion affects only the conformal energy of the mapping $S\rightarrow S^H$. However, circumferential updates may produce a large number of elongated triangles, which can lead to numerical instability when evaluating the conformal energy derivation. Moreover, optimizing the circumferential motion has little influence on the final toolpath. Therefore, only radial updates of $V_i^H$ are considered. A practical way to enforce Constraints 1 and 2 is as follows.

\textbf{Constraint 1}. Let $N^1(V_i^H)$ denote the one-ring neighborhood of the mesh vertex $V_i^H$. For each triangle in $N^1(V_i^H)$, consider the two angles opposite $V_i^H$. The angle bisectors on the side containing $V_i^H$ form a convex region, denoted by $CH(V_i^H)$, whose boundary is written as $\partial CH(V_i^H)$; see Figures~\ref{fig:meshInitialization}(c) and (d). If $V_i^H$ moves within the open region $CH(V_i^H)\setminus\partial CH(V_i^H)$ during each iteration, no triangle inversion will occur.

\textbf{Constraint 2}. During each iteration, all vertices on the same boundary $\Gamma_i^H$ must undergo the same radial displacement.

\subsubsection{Variational Optimization of Level Set Trajectories under Iterative Constraints}

Our objective is to determine $T_{opt}$ in Equation~\ref{eq:itertarget}. As described in Section 2.3.1, this problem can be transformed into finding the optimal positions of the vertex set $V^H$ on the discrete mesh $S^H$.

The scalar value $T(V)$ corresponding to each vertex $V$ on the discrete mesh $S$ is given by:
\begin{equation}
T(V) = \| V^H \|
\end{equation}

The scalar values at other locations on $S$ are obtained using triangular barycentric interpolation. Let $N^2(p)$ denote the local mesh formed by the two-ring neighbors of node $p$, where $N^2(p)\in S$. The local metric value denoted as $E(N^2(p))$ corresponding to this mesh can still be computed using Equation~\ref{eq:Eall}. For any vertex $V^H_i\in V^H$, when it moves by a small distance $\partial V^H_i$ along the radial direction $\overrightarrow{V^H_i}$, only the energy on the local mesh $N^2(V^H_i)$ is affected. Hence,
\begin{equation}
\frac{\partial E}{\partial V_i^H} = \frac{\partial E\left( N^2\left( V_i^H \right) \right)}{\partial V_i^H}
\label{eq:partialNode}
\end{equation}
where $\frac{\partial E}{\partial V^H_i}$ indicates whether moving $V^H_i$ along the radial direction $\overrightarrow{V^H_i}$, either forward or backward, will decrease $E$. Owing to Constraint 2, all vertices on the same boundary $\Gamma^H_i$ should move synchronously, either collectively towards or away from the circle center. The corresponding variation of $E$ caused by the synchronous iterative movement of boundary $\Gamma^H_i$ can be expressed as:
\begin{equation}
\frac{\partial E}{\partial \Gamma_i^H} = \sum_{j \in |\Gamma_i^H|} \frac{\partial E}{\partial V_j^H},\ \text{where}\ i = 1,2,\dots,m
\label{eq:partialEdge}
\end{equation}

By combining Equations~\ref{eq:partialNode} and \ref{eq:partialEdge}, the gradient set $G^H$ can be expressed as:

\begin{equation}
\begin{aligned}
G^H
={}&
\left\{
\frac{\partial E}{\partial V_i^H}
\,\middle|\,
i \in \lvert V^H \setminus \Gamma^H \rvert
\right\}
\\
&\cup
\left\{
\frac{\partial E}{\partial \Gamma_j^H}
\,\middle|\,
j = 1,2,\dots,m
\right\}
\end{aligned}
\end{equation}

As illustrated in Figures~\ref{fig:meshInitialization}(c) and (d), for a vertex $V^H_i\in V^H \setminus \Gamma^H$ that is not located on the boundary $\Gamma^H$, the corresponding convex hull $CH(V^H_i)$ is bounded. For a vertex $V^H_i\in \Gamma^H$ located on the boundary, the corresponding convex hull $CH(V^H_i)$ may be either bounded or unbounded. However, introducing a circular-arc boundary to $CH(V^H_i)$ (Figure~\ref{fig:meshInitialization}(d)) ensures that it becomes bounded, thereby preventing the excessive displacement of boundary vertices during iteration and improving numerical stability. For simplicity, the convex hull with an added circular-arc boundary is denoted as $CH(V^H_i)$.

The meridian line passing through $V^H_i$ intersects the boundary $\partial CH(V^H_i)$ at two points. The point farther from the center $O^H$ is denoted as $V^H_{i-1}$, and the other point is denoted as $V^H_{i-2}$. Accordingly, we define $\lambda_i^1 = \left\| V_i^H V_{i-1}^H \right\|$ and $\lambda_i^2 = -\left\| V_i^H V_{i-2}^H \right\|$. To prevent mesh inversion, the iteration of each $V^H_i\in V^H \setminus \Gamma^H$ should satisfy:
\begin{equation}
V_i^H = V_i^H + \lambda_i \frac{V_i^H}{\|V_i^H\|},\ \text{where}\ \lambda_i \in \left( \lambda_i^2, \lambda_i^1 \right),\ i \in |V^H \setminus \Gamma^H|
\label{eq:nodelabda}
\end{equation}

For the set of vertices on the boundary $\Gamma^H_i$, where $i=1,2,\ldots,m$, their synchronous iteration is subject to a more stringent constraint:
\begin{equation}
\Gamma_i^H = \Gamma_i^H + \lambda_j \frac{\Gamma_i^H}{\|\Gamma_i^H\|},\\ 
\label{eq:edgelabda}
\end{equation}
where
\begin{equation}
 \lambda_j \in \left( \lambda_j^2 = \max\left( \lambda_{|\Gamma_i^H|}^2 \right),\ \lambda_j^1 = \min\left( \lambda_{|\Gamma_i^H|}^1 \right) \right)
 \label{eq:edgelabdaconstrain}
\end{equation}

Here, $\lambda_{|\Gamma_i^H|}^1$ and $\lambda_{|\Gamma_i^H|}^2$ denote the sets of $\lambda^1_j$ and $\lambda^2_j$, respectively, for all $j\in |\Gamma^H_i|$.

For any $G^H_i$ with $i\in |G^H|$, a corresponding feasible interval $[\lambda^1_i,\lambda^2_i]$ can be obtained by combining Equations~\ref{eq:nodelabda} and \ref{eq:edgelabda}. The gradient set $G^H$ is then normalized within feasible intervals using the following procedure:
\begin{equation}
\begin{cases}
G^{H1} = \left\{ \frac{-G_i^H}{\left\| (1-I_i)\lambda_i^1 + I_i\lambda_i^{2} \right\|},\ i \in |G^H| \right\} \\
I_i = 0\ \text{for}\ G_i^H < 0,\ I_i = 1\ \text{for}\ G_i^H > 0 \\
G^{H2} = \log\left( \frac{1}{C} - 1 \right) \frac{G^{H1}}{\max\left( \|G^{H1}\| \right)},\ C \in (0.5, 1) \\
\lambda = \left( \frac{2}{1+e^{G^{H2}}} - 1 \right) \left\| (1-I_i)\lambda_i^1 + I_i\lambda_i^2 \right\|
\end{cases}
\label{eq:softmax}
\end{equation}
where $G^{H1}$ represents the ratio between the energy gradient and the feasible movement range, and $C\in (0.5,1)$ is a constant. Larger values of $C$ lead to faster but less accurate convergence, whereas a value of 0.9 has been observed to perform well across all test cases. By substituting $\lambda$ from Equation~\ref{eq:softmax} into Equations~\ref{eq:nodelabda} and \ref{eq:edgelabda}, a variational iteration process satisfying the required constraints can be obtained.

\section{Experiment and analysis}
\subsection{Numerical experiments}

\subsubsection{Effectiveness analysis of mesh initialization}
\begin{figure*}
  \centering
  \includegraphics[width=0.8\linewidth]{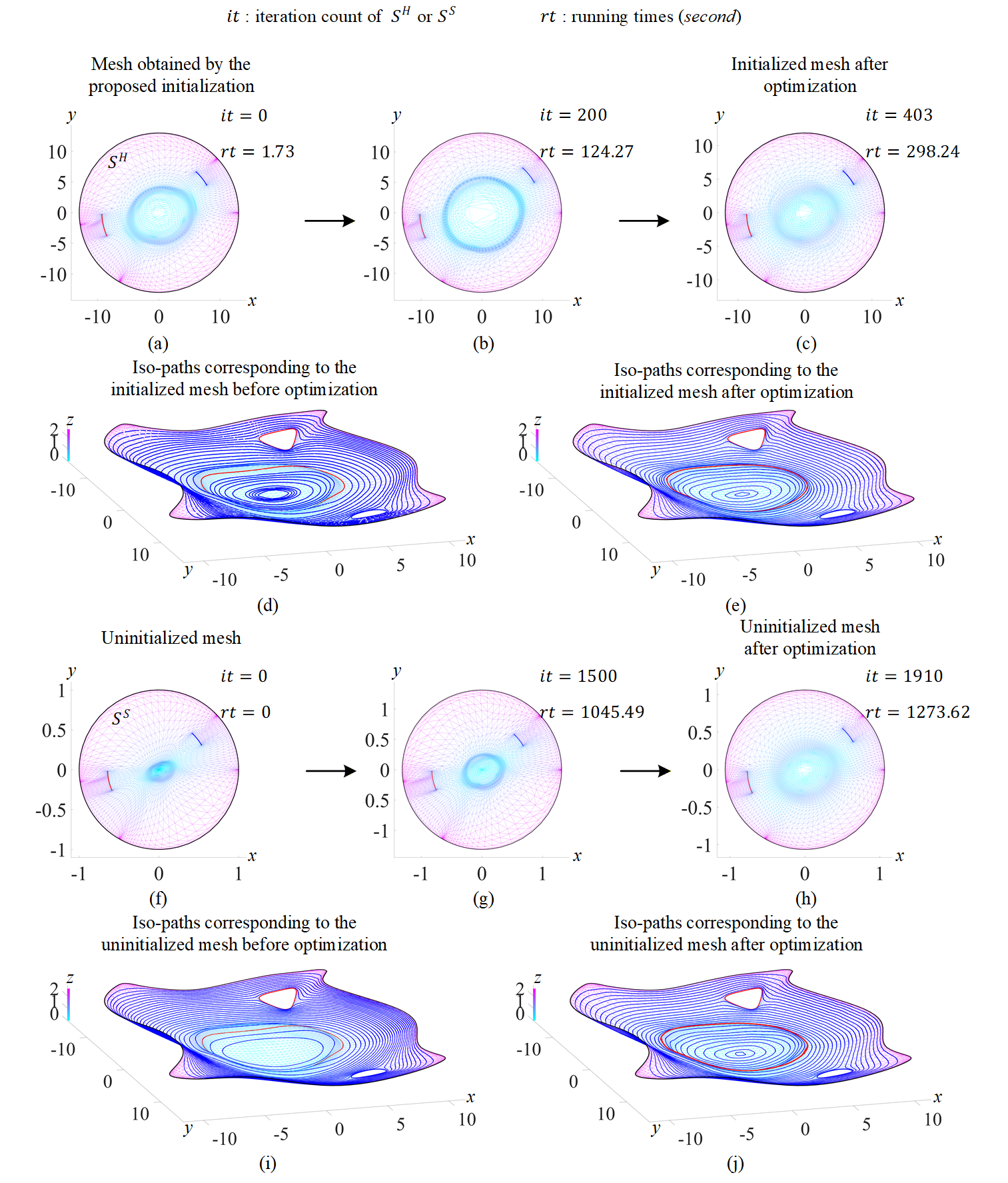}
  \caption{\label{fig:simulationComparision}Variational iteration process of the mesh. (a) $S^H$ mesh obtained from the initialization of $S^S$; (b) $S^H$ mesh during iteration; (c) converged $S^H$ mesh; (d) iso-paths of the initialized $S^H$ mesh; (e) iso-paths of the converged $S^H$ mesh; (f) uninitialized $S^S$ mesh; (g) $S^S$ mesh during iteration; (h) converged $S^S$ mesh; (i) iso-paths of the uninitialized $S^S$ mesh; (j) iso-paths of the converged $S^S$ mesh.}
\end{figure*}

The algorithm was run on a desktop with an Intel Core i5-10400F CPU and an Nvidia 1600 GPU. Figures~\ref{fig:simulationComparision}(a)–(c) and Figures~\ref{fig:simulationComparision}(f)–(h) show the mesh iteration processes starting from initialized $S^H$  and uninitialized $S^S$, respectively, whereas Figures~\ref{fig:simulationComparision}(d), (i), (e), and (j) illustrate the corresponding iso-paths before and after iterations. By comparing Figure~\ref{fig:simulationComparision}(e) with (j), the converged trajectories are nearly identical. However, using the conformal-slit-mapping initialization reduced the convergence time by 76.58\%, from 1273.62 s to 298.24 s. This indicated that the proposed initialization effectively reduced both the number of iterations and the overall computation time without introducing any noticeable influence on the final trajectory. The underlying reasons could be as follows:
\begin{itemize}
\item Although the trajectories corresponding to the isoparametric curves of $S^H$ and $S^S$ exhibited similar overall shapes, the initial states shown in Figures~\ref{fig:simulationComparision}(a) and (b) revealed an evident difference. In particular, in regions near the center, where conformal mapping introduced severe radial distance distortion \cite{Xu2019Contour}, the trajectory spacing generated from the  $S^H$ isoparametric curves was noticeably more uniform than that generated from $S^S$. As a result, fewer iterations are required to achieve uniform spacing during the optimization process.
\item The key step in transforming $S^S$ into $S^H$ was the $f_{\min}$ operation defined in Equation~\ref{eq:fconstraint}, whose computational complexity was $O\left( N_F \log\left( \frac{1}{E_\varepsilon^S} \right) \right)$, where $N_F$ denotes the number of triangular faces of $S$ and $E^S_{\varepsilon}$ is the convergence energy threshold. This procedure has been demonstrated to be efficient in a previous study \cite{Shen2025Conformal} and requires only 1.73 s in the test case. Therefore, it is reasonable to invest a small amount of additional computation time in mesh initialization to achieve a substantial reduction in the overall optimization costs.
\end{itemize}

Notably, a difference was observed between the red trajectories in Figures~\ref{fig:simulationComparision}(d) and (e). The optimized trajectory aligned more closely with the low-curvature direction of the boss feature than the trajectory generated directly from conformal slit mapping. This alignment can help avoid frequent crossings over the boss, thereby reducing abrupt directional changes in the toolpath.

\subsubsection{Effect of the weighting parameter $\alpha$}
\begin{figure*}
  \centering
  \includegraphics[width=0.8\linewidth]{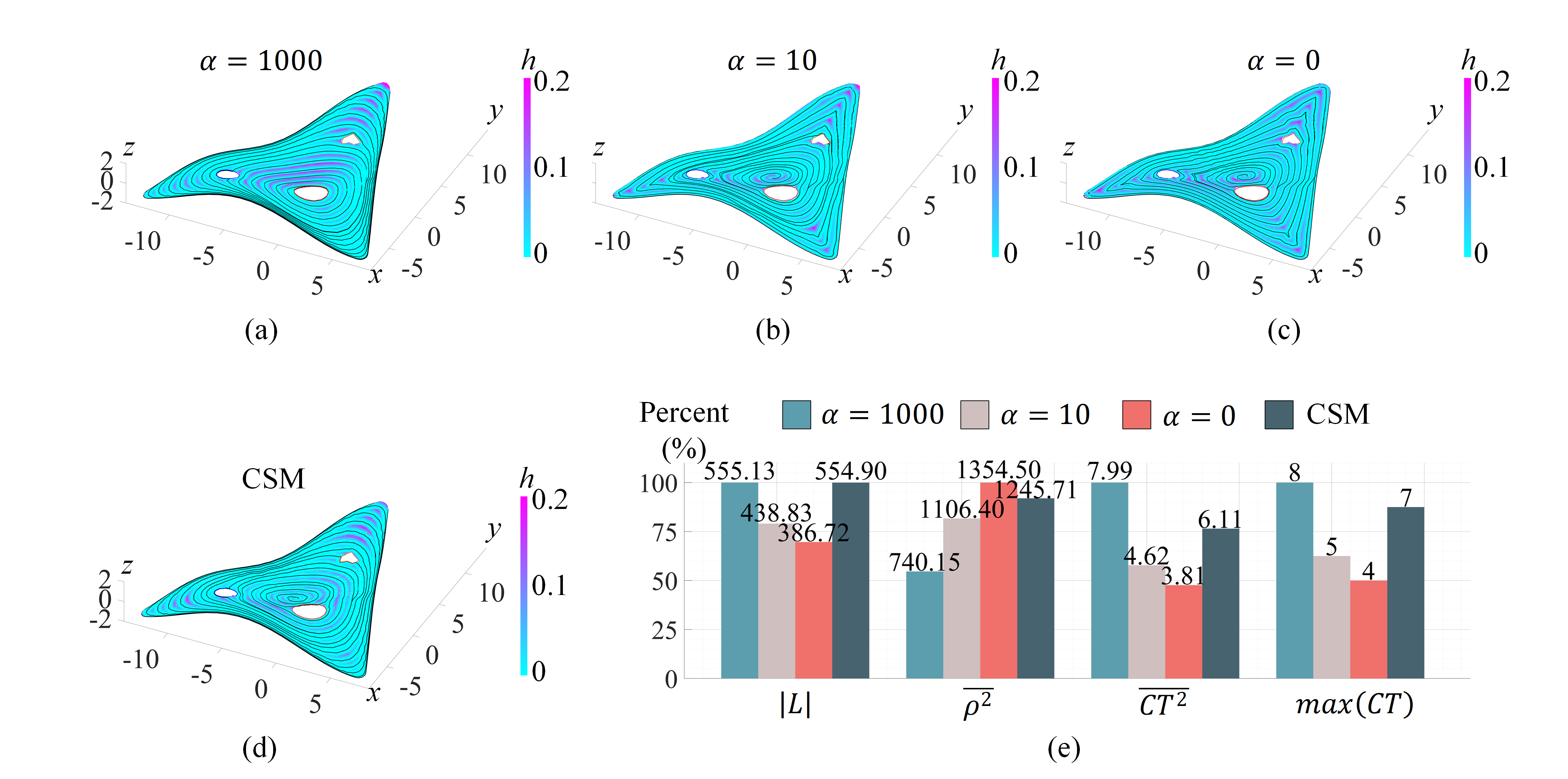}
  \caption{\label{fig:evaluationmetrics}Tool contact trajectories and evaluation metrics: (a)–(c) effect of weighting parameter $\alpha$; (d) conformal slit mapping method (reference); (e) multi-indicator comparison.}
\end{figure*}

Figures~\ref{fig:evaluationmetrics}(a)–(c) illustrate the differences among tool contact trajectories generated using different values of the weighting parameter $\alpha$ in Equation~\ref{eq:Eall}, with the prescribed machining scallop height $h_{set}$ fixed at 0.2. Figure~\ref{fig:evaluationmetrics}(d) shows the trajectory generated using the conformal slit mapping method \cite{Shen2025Conformal}, with the origin mapped point $\Theta$ set to be the same as that in Figure~\ref{fig:evaluationmetrics}(b); this result was used as the reference. Figure~\ref{fig:evaluationmetrics}(e) compares the generated tool contact trajectories in terms of total path length $|L|$, turning smoothness index $\overline{\rho^2}$, machining redundancy index $\overline{CT^2}$, and the maximum repetition coverage count $\max(CT)$. The trajectory smoothness index $\overline{\rho^2}$ is defined as:
\begin{equation}
\overline{\rho^2} = \int_L \rho^2 dL
\end{equation}
where $dL$ represents the differential element of the tool-contact trajectory $L$, and $\rho$ is the curvature corresponding to each differential segment. The squared curvature term $\rho^2$ imposes a larger penalty on high-curvature regions, and a smaller value of the smoothness metric $\overline{\rho^2}$ indicates a smoother trajectory.

\begin{figure}
  \centering
  \includegraphics[width=0.8\linewidth]{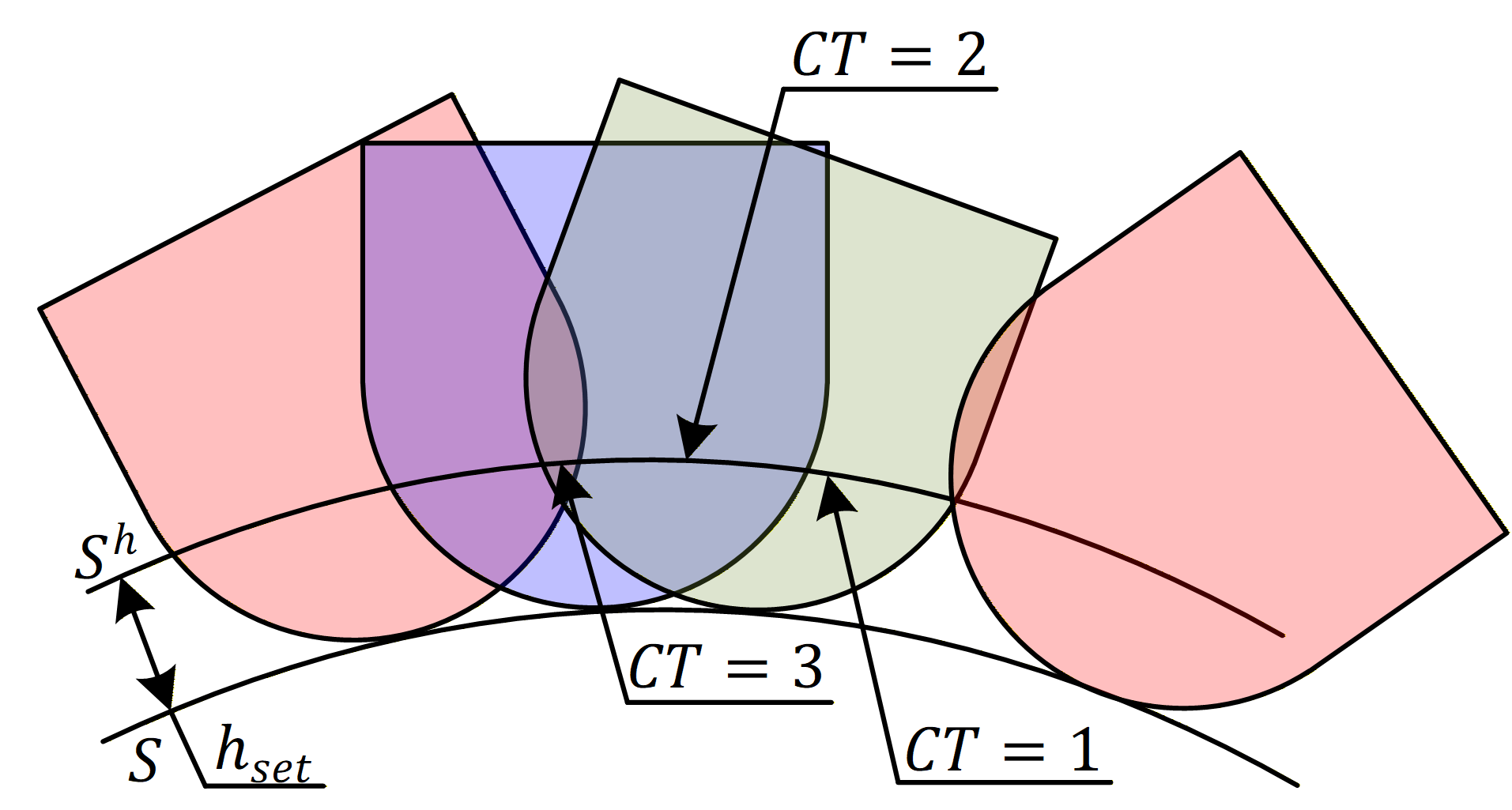}
  \caption{\label{fig:crosssection}Cross-sectional view of the coverage of iso-scallop-height surfaces by adjacent tool positions.}
\end{figure}

The machining redundancy index $\overline{CT^2}$ is defined as follows:
\begin{equation}
\overline{CT^2} = \int_{S^h} CT^2 ds
\end{equation}
where $S^h$ denotes the iso-scallop-height surface obtained by offsetting the original surface $S$ along the normal direction by the prescribed scallop height $h_{set}$,  $ds$ represents a differential element of $S^h$, and $CT$ indicates the number of times a differential element is covered by different trajectories (Figure~\ref{fig:crosssection}). By adopting the squared term of $CT$, regions with multiple redundant coverage were penalized more heavily. A smaller value of the redundancy metric $\overline{CT^2}$ indicated less redundant machining, and also, a smaller maximum repetition count $\max(CT)$ corresponded to reduced trajectory redundancy.

As shown in Figure~\ref{fig:evaluationmetrics}(e), as the smoothness weighting parameter $\alpha$ decreased, the metrics $|L|$, $\overline{CT^2}$, and $\max(CT)$ decreased simultaneously, demonstrating the reduced trajectory redundancy as well as more consistent spacing and scallop height. Furthermore, the smoothness metric  $\overline{\rho^2}$ increased, implying a reduction in trajectory smoothness. Therefore, $\alpha$ can be adjusted according to specific requirements, where a smaller $\alpha$ favors shorter trajectories with more uniform scallop height, whereas a larger $\alpha$ favors trajectories with smoother directional transitions. When $\alpha=10$, the resulting trajectory achieved uniformly better values across all metrics than the trajectory generated using conformal slit mapping, further confirming the effectiveness of the proposed method.

\subsubsection{Generality, Limitations, and Insights into Subregion Decomposition}
\begin{figure*}
  \centering
  \includegraphics[width=0.8\linewidth]{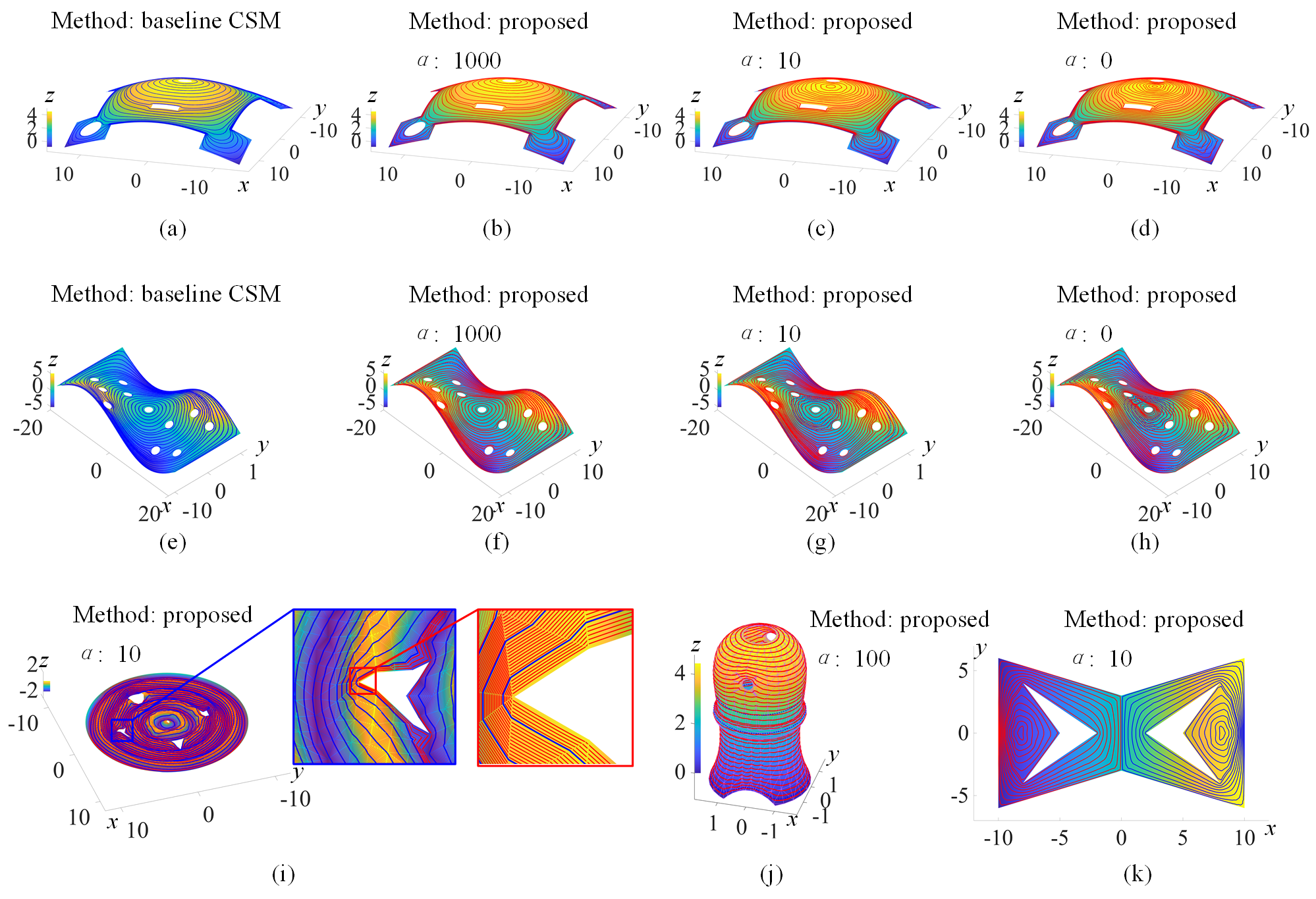}
  \caption{\label{fig:generality_limitations}Applicability of the proposed method to representative complex freeform surfaces, its limitation on subregions with large width variation, and its implications for future subregion decomposition.}
\end{figure*}

Figure~\ref{fig:generality_limitations} further demonstrates the applicability of the proposed method to several representative complex freeform surfaces. Figure~\ref{fig:generality_limitations}(a) shows the baseline trajectory generated by the CSM-based method, whereas Figures~\ref{fig:generality_limitations}(b)–(d) present the trajectories generated by the proposed method under different values of $\alpha$. The blue dashed curves indicate the reference trajectories for comparison. Similarly, Figure~\ref{fig:generality_limitations}(e) shows another baseline trajectory generated by the CSM-based method, whereas Figures~\ref{fig:generality_limitations}(f)–(h) show the corresponding trajectories generated by the proposed method under different values of $\alpha$. For these challenging surfaces, conventional toolpath planning methods often suffer from discontinuities, self-intersections, or boundary crossing. In contrast, the trajectories generated by the proposed method remain continuous, boundary-conforming, and free of self-intersections over a range of $\alpha$ values.

Figure~\ref{fig:generality_limitations}(i), together with its enlarged local view, further illustrates the high-density trajectories generated by the proposed method on a multiply connected surface for finishing applications. The blue solid curve represents a portion of the generated trajectory and is highlighted to show that the overall trajectory remains globally non-self-intersecting and does not cross any boundary component even under dense path arrangements.

Figure~\ref{fig:generality_limitations}(j) illustrates the trajectory generated by the proposed method on a deep-cavity structure, while Figure~\ref{fig:generality_limitations}(k) highlights the potential of integrating the proposed method with existing cell decomposition strategies \cite{patel2017decomposition,abrahamsen2019cuthole,Sun2016Smooth}. When the local width of a surface region varies substantially, covering the entire region with a single spiral trajectory may lead to pronounced spacing variation and, consequently, severe scallop-height inconsistency. However, this issue can be significantly mitigated through appropriate subregion decomposition. As shown in Figure~\ref{fig:generality_limitations}(k), the proposed method further reveals, in contrast to previous studies, that it is unnecessary for every decomposed subregion to be completely hole-free. Some holes that do not significantly affect scallop-height consistency may be retained, thereby reducing bridging while still generating trajectories that are shorter, smoother, more scallop-height-consistent, and free of self-intersections.

\subsection{Experiment on surface milling}

\begin{figure}
  \centering
  \includegraphics[width=0.8\linewidth]{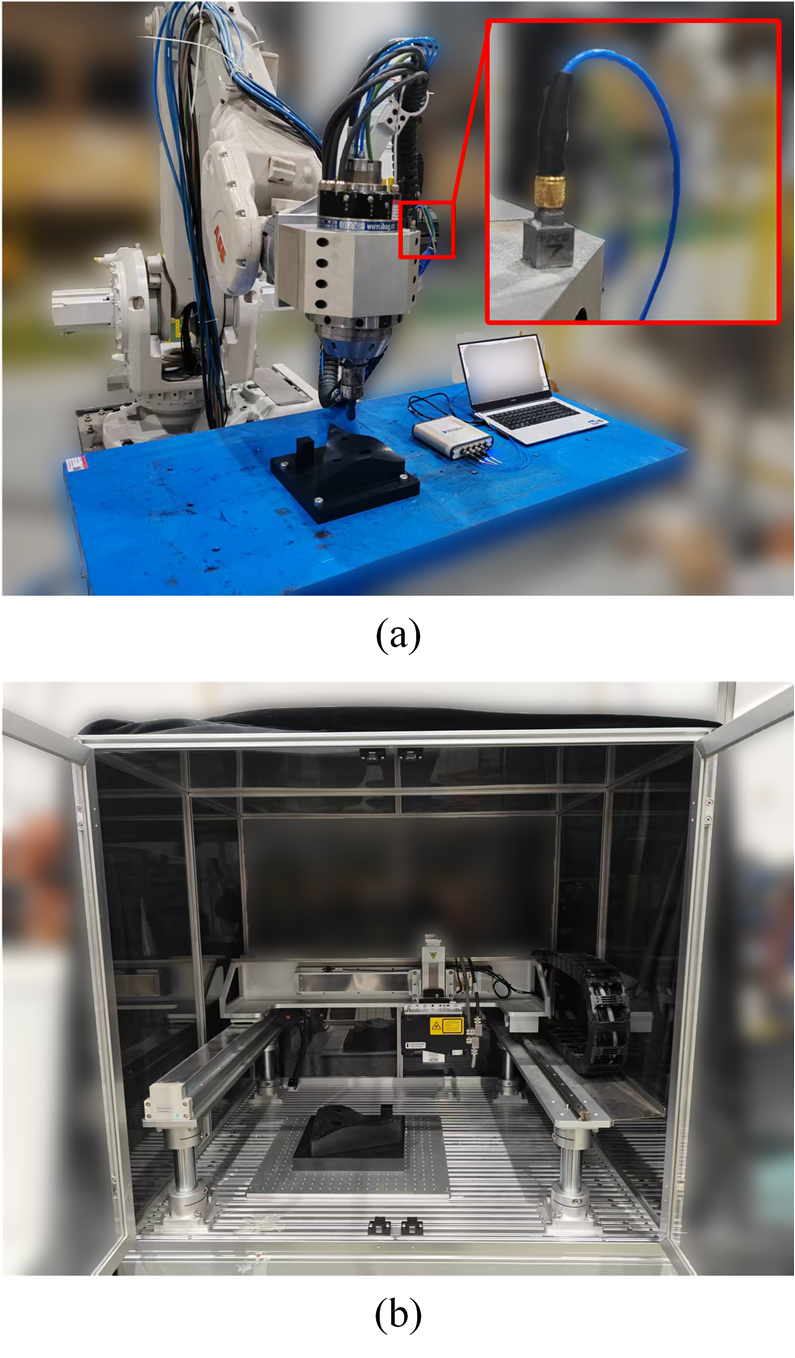}
  \caption{\label{fig:platform}Experimental setup. (a) IRB6600 robotic milling system equipped with a PCB triaxial accelerometer and (b) three-coordinate laser measurement system.}
\end{figure}

An experimental comparison was conducted to evaluate the machining performance of two trajectories: Path 1 generated by the proposed algorithm with $\alpha=10$, and Path 2 generated using the conformal slit mapping method \cite{Shen2025Conformal}. Figure~\ref{fig:platform}(a) shows the IRB6600 robotic milling system, in which a PCB triaxial accelerometer was mounted on the spindle to measure the dynamic cutting vibrations at a sampling frequency of 2000 Hz. Figure~\ref{fig:platform}(b) shows the three-coordinate laser measurement system for surface inspection, which provided a measurement accuracy of 20\,$\mu$\text{m}.

All non-critical control variables are summarized in Table~\ref{tab:performancemetrics} and were kept constant to ensure a fair and unbiased comparison.
\begin{table*}
  \caption{Machining performance metrics under identical control conditions.}
  \label{tab:performancemetrics}
 \resizebox{\textwidth}{!}{ \begin{tabular}{cccccccl}
    \toprule
Variables&	Tool radius (mm)&	Number of flutes&	Tool overhang length (mm)&	Spiral motion direction&	Axial depth of cut (mm)&	Spindle speed (rpm)	&Workpiece material\\
    \midrule
Values&	10	&2	&79.2&	From inside to outside	&4.00&	1000	&Engineering plastic (ABS)\\
  \bottomrule
\end{tabular}}
\end{table*}

\begin{figure*}
  \centering
  \includegraphics[width=0.8\linewidth]{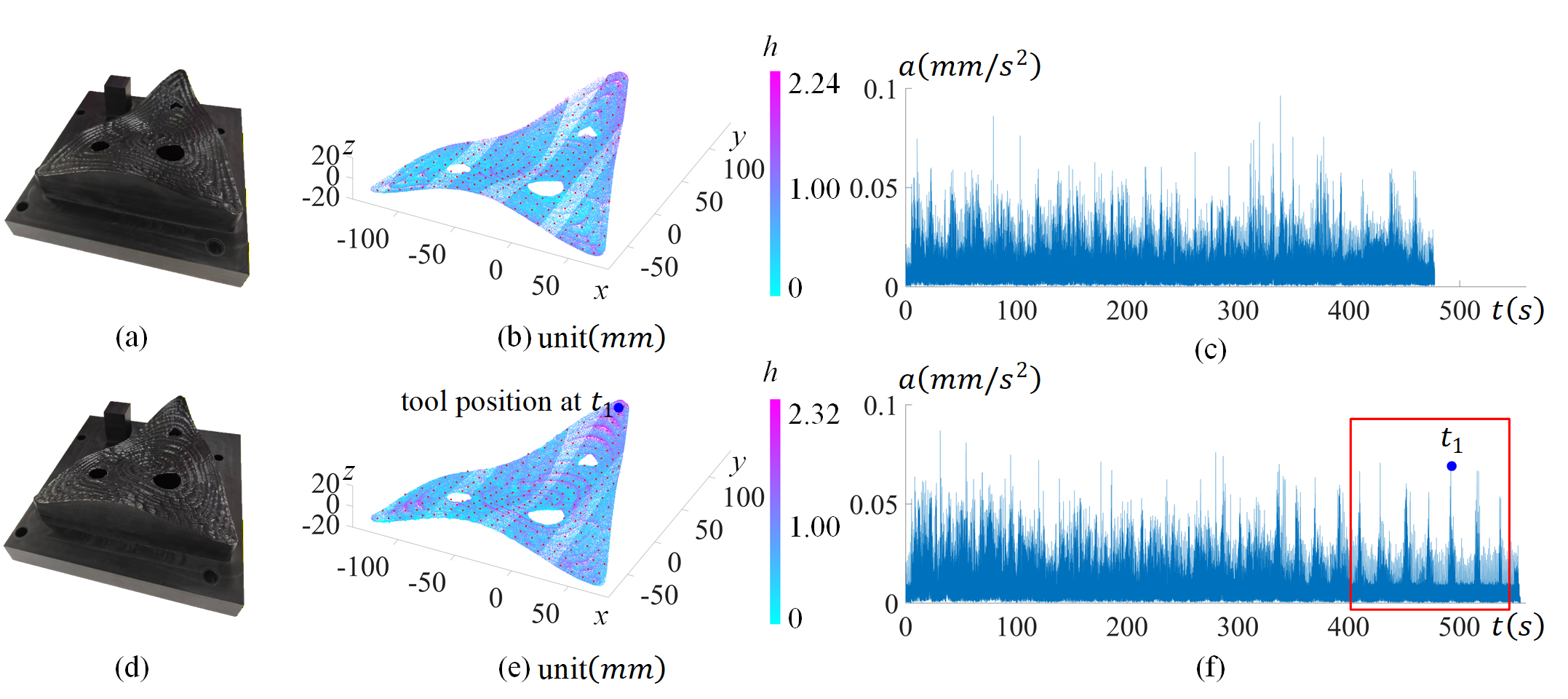}
  \caption{\label{fig:experimentalresult}Comparison of experimental results (upper: proposed method; lower: conformal slit mapping), showing (from left to right) machined workpieces, scanned point clouds, and spindle acceleration signals.}
\end{figure*}

Figure~\ref{fig:experimentalresult} presents the surface-scanned point clouds and measured spindle acceleration signals from left to right for the machined workpieces, where the upper and lower rows correspond to the experimental results for Path 1 and Path 2, respectively. Except for the prescribed feed rate, all machining parameters were kept identical in the two experiments. As shown in Figures~\ref{fig:experimentalresult}(b) and (e), the machining scallops of Path 1 and Path 2 closely match the simulation results in Figures~\ref{fig:evaluationmetrics}(b) and (d). The maximum machining residuals were 2.24 mm and 2.32 mm (corresponding to deviations of 12.0\% and 16.0\%, respectively, from the set value of 2 mm). However, the difference in the maximum residuals between the two experiments is less than 5\%, and the subsequent measurements exhibit clear and consistent trends, supporting the validity of the experimental results.

Table~\ref{tab:quantitativecomparison} summarizes the quantitative comparison of machining performance metrics, with the third row presenting the ratio of the metrics for Path 1 to Path 2. Specifically, $F_R$ denotes the prescribed feed rate, and $time$ represents the total machining time; and $S_c$ is the global scallop height consistency metric defined based on practical part quality inspection procedures. A set of sampling points $Pt$ was selected on the machined surface, as indicated by the red points in Figures~\ref{fig:experimentalresult}(b) and (e). For each point $Pt_i\in Pt$, the maximum scallop height $msh(Pt_i)$ was evaluated within a local neighborhood defined as a spherical region centered at $Pt_i$ with tool radius $K_c$. The metric $S_c$ was defined as the variance of $msh(Pt_i)$ over all sampling points:
\begin{equation}
S_c = \text{var}\left( \left\{ Pt_i \in P_t \mid msh(Pt_i) \right\} \right)
\end{equation}

\begin{table*}
  \caption{\label{tab:quantitativecomparison}Quantitative comparison of machining performance metrics between Path 1 (proposed method) and Path 2 (conformal slit mapping).}
 \resizebox{\textwidth}{!}{\begin{tabular}{ccccccccl}
    \toprule
&	$F_R$ $mm/s$ &	$Time\ sec$ &	$S_c\ mm^2$ (experiment) &	$S_c\ mm^2$ (simulation) &	$a_{mean}\ 10^{-3}mm/s^2$	& $a_{var}\ 10^{-5}mm^2/s^4$ & $a_{cu}\ 10^{-1}mm^2/s^3$ & $a_{tcu}\ 10^{-4}mm^2/s^3$\\
    \midrule
Path 1&	10.00&	466.33&	0.2101&	0.2007	&9.64&	3.53&	6.03&	6.51\\
Path 2&	11.00&	543.75&	0.2228&	0.2140	&9.21&	3.98	&6.85	&8.38\\
Ratio&90.91\%&	85.76\%	&94.30\%&	93.79\%	&104.67\%&	88.69\%&	87.93\%	&77.38\%\\
  \bottomrule
\end{tabular}}
\end{table*}

In addition, $a_{mean}$ and $a_{var}$ denote the mean value and variance of the measured spindle acceleration signal, respectively. The cumulative impact indicator $a_{cu}$ and the threshold-based cumulative impact indicator $a_{tcu}$, defined as the accumulated acceleration exceeding a threshold of $0.05 mm/s^2$, were employed to evaluate the overall and extreme impact loads experienced by the actuator throughout the machining process. They are defined as follows:
\begin{equation}
a_{cu} = \int a^2 dt
\end{equation}
\begin{equation}
a_{tcu} = \int \text{exceed}_{T_{0.05}}(a^2) dt
\end{equation}
where
\begin{equation}
\text{exceed}_{T_{0.05}}(a^2) = 
\begin{cases} 
a^2 & \text{for } a > 0.05\,\text{mm/s}^2, \\
0 & \text{otherwise}
\end{cases}
\end{equation}

The experimental results provide insights into the causal relationships among trajectory characteristics, machining efficiency, scallop height consistency, and process stability. Although both trajectories share identical topological properties, including a single tool entry and exit, an outward spiral motion, and the absence of hole crossing or self-intersection, their geometric organization over the surface leads to substantially different machining responses.

Although the prescribed feed rate of Path 1 ($F_R=10 mm/s$) was 9.09\% lower than that of Path 2 ($F_R=11 mm/s$), Path 1 achieved a 14.24\% shorter machining time (466.33 s versus 543.75 s). This improvement was primarily attributed to a 20.92\% reduction in the path length resulting from the elimination of redundant coverage by the proposed algorithm.

The 5.70\% reduction in $S_c$ obtained for Path 1 demonstrated a more uniform scallop height distribution over the surface, which was consistent with the 6.21\% reduction predicted by simulation. This improvement was mainly due to the ability of the proposed method to enforce global regularity in the scallop height distribution along the trajectory.

From a dynamic perspective, Path 1 exhibited a 4.67\% increase in the mean spindle acceleration, reflecting a higher average material removal rate. However, machining stability was governed not by the absolute cutting load but by its variation along the trajectory. The consistently lower values of $a_{var}$, $a_{cu}$, and $a_{tcu}$ observed for Path 1 indicated a pronounced reduction in impulsive load fluctuations, with all three indicators reduced by more than 10\% (approximately 11\%, 12\%, and 23\%, respectively) compared with Path 2. This behavior can be attributed to the smaller variation in the material removal rate along Path 1.

This effect was particularly evident in the red-highlighted region of Figure~\ref{fig:experimentalresult}(f), where the toolpath approached the outer boundary of the workpiece. In this area, the trajectory generated by conformal slit mapping exhibited pronounced acceleration pulses, resulting from abrupt variations in local cutting engagement, especially in corner-dominated regions, where the cutting rate became significantly higher than in adjacent areas. In contrast, the smoother spatial distribution of Path 1 alleviated abrupt engagement transitions, leading to a more uniform and stable dynamic response.

Overall, the experimental results suggest that regulating the trajectory redundancy and local engagement variation through the weighting parameter enables concurrent improvements in machining efficiency, scallop height consistency, and process stability. This unified control mechanism highlights the practical advantages of the proposed trajectory generation method over conformal slit mapping-based approaches for complex freeform surface milling.


\section{Summary and outlook}
\subsection{Summary}
This study presented a variational scalar-field framework for multiply connected (porous) freeform surfaces with the aim of achieving two core objectives: (1) the generation of boundary-conforming, continuous, and self-intersection-free spiral toolpaths through topology-preserving mesh deformation, and (2) the simultaneous optimization of global scallop height uniformity and path smoothness via scalar-field variational optimization, in which an adjustable weighting parameter was introduced to balance these competing objectives. Compared with conformal slit mapping, the proposed approach improved the machining efficiency by 14.24\%, enhanced the scallop height uniformity by 5.70\%, and reduced the milling impact by more than 10\%.

To guarantee boundary conformity and avoid zero-gradient singularities, the scalar field must satisfy strict interior and boundary constraints. These constraints are reformulated as topology-preserving mesh optimization and boundary co-circularity enforcement, ensuring their preservation throughout the entire iterative optimization process and effectively preventing boundary nonconformity, discontinuities, and self-intersection of toolpaths.

Furthermore, conformal slit mapping was employed to construct a feasible initial mesh configuration that inherently satisfied both types of constraints. This initialization mitigated numerical instability, suppressed radial distortion induced by conformal mapping, and significantly reduced the required number of iterations and overall computational cost, thereby providing a stable and efficient foundation for subsequent scalar-field optimization.

\subsection{Outlook}
Because the scalar field can be influenced by both its initialization and locally linearized perturbation updates, attaining a globally optimal solution remains challenging. Introducing stochastic exploration into the iterative optimization process may provide a viable means of approaching more globally optimal scalar-field configurations.

In addition, the Riemannian metric derived from scallop height analysis can be replaced with metrics tailored to other domain-specific coverage objectives \cite{Zou2021Length}, such as additive manufacturing or UAV area coverage. Consequently, the proposed framework exhibits strong potential for extension to a broader class of coverage path planning problems.

\section{Acknowledgement}
This study was supported by the National Natural Science Foundation of China (52188102, 52375495). We sincerely thank Dr. Yang Zeyuan for valuable comments and insightful suggestions that greatly improved the quality of this manuscript. 

\bibliographystyle{cas-model2-names}

\bibliography{cas-refs}

@ARTICLE{Acar2003Path,
  author = {Acar, Ercan U. and Choset, Howie and Zhang, Yangang and Schervish, Mark},
  title = {Path Planning for Robotic Demining: Robust Sensor-Based Coverage of Unstructured Environments and Probabilistic Methods},
  journal = {Int. J. Robot. Res.},
  year = {2003},
  volume = {22},
  number = {7-8},
  pages = {441-466},
  doi = {10.1177/02783649030227002}
}

@inproceedings{Ban2013Topology,
  author = {Ban, Xiaomeng and Goswami, Mayank and Zeng, Wei and Gu, Xianfeng and Gao, Jie},
  title = {Topology dependent space filling curves for sensor networks and applications},
  booktitle = {2013 Proceedings IEEE INFOCOM},
  year = {2013},
  publisher = {IEEE},
  address = {Turin, Italy},
  pages = {14-19},
  doi = {10.1109/INFCOM.2013.6567019}
}

@ARTICLE{Bartoň2021Geometry,
  author = {Bartoň, Michael and Bizzarri, Michal and Rist, Florian and Sliusarenko, Oleksii and Pottmann, Helmut},
  title = {Geometry and tool motion planning for curvature adapted CNC machining},
  journal = {ACM Trans. Graph.},
  year = {2021},
  volume = {40},
  number = {4},
  pages = {1-16},
  doi = {10.1145/3450626.3459837}
}

@ARTICLE{Bohez2000Adaptive,
  author = {Bohez, E. and Makhanov, S. S. and Sonthipermpoon, K.},
  title = {Adaptive nonlinear tool path optimization for five-axis machining},
  journal = {Int. J. Prod. Res.},
  year = {2000},
  volume = {38},
  number = {17},
  pages = {4329-4343},
  doi = {10.1080/00207540050205127}
}

@ARTICLE{Campen2016Bijective,
  author = {Campen, Marcel and Silva, Cláudio T. and Zorin, Denis},
  title = {Bijective maps from simplicial foliations},
  journal = {ACM Trans. Graph.},
  year = {2016},
  volume = {35},
  number = {4},
  pages = {1-15},
  doi = {10.1145/2897824.2925890}
}

@inproceedings{Chen2025Singularity,
  author = {Chen, Haokun and He, Dong and Hao, Jiancheng and Hu, Pengcheng and Tang, Kai},
  title = {Singularity-free PINN-based path planning method for the table-tilt five-axis machine*},
  booktitle = {2025 IEEE International Conference on Real-time Computing and Robotics (RCAR)},
  year = {2025},
  publisher = {IEEE},
  address = {Toyama, Japan},
  pages = {636-641},
  doi = {10.1109/RCAR65431.2025.11139420}
}

@ARTICLE{Chen2021Investigation,
  author = {Chen, Zhongwei and Wu, Xian and Zeng, Kai and Shen, Jianyun and Jiang, Feng and Liu, Zhongyuan and Luo, Wenjun},
  title = {Investigation on the Exit Burr Formation in Micro Milling},
  journal = {Micromachines (Basel)},
  year = {2021},
  volume = {12},
  number = {8},
  pages = {952},
  doi = {10.3390/mi12080952}
}

@ARTICLE{Choi2021Efficient,
  author = {Choi, Gary P. T.},
  title = {Efficient Conformal Parameterization of Multiply-Connected Surfaces Using Quasi-Conformal Theory},
  journal = {J. Sci. Comput.},
  year = {2021},
  volume = {87},
  pages = {70},
  doi = {10.1007/s10915-021-01479-y}
}

@ARTICLE{Della2023Discontinuous,
  author = {Della Santa, Francesco and Pieraccini, Sandra},
  title = {Discontinuous neural networks and discontinuity learning},
  journal = {J. Comput. Appl. Math.},
  year = {2023},
  volume = {419},
  pages = {114678},
  doi = {10.1016/j.cam.2022.114678}
}

@ARTICLE{Pereira2025EnhancingPO,
	title={Enhancing productivity of helical milling of Inconel 718 by optimization with constraint learning},
	author={Robson Bruno Dutra Pereira and Gaizka G{\'o}mez-Escudero and Amaia Calleja-Ochoa and Haizea Gonz{\'a}lez-Barrio and Carlos Henrique Lauro and Lincoln Cardoso Brand{\~a}o and Luis Norberto L{\'o}pez de Lacalle},
	journal={The International Journal of Advanced Manufacturing Technology},
	year={2025},
	volume={138},
	pages={783 - 804},
	url={https://api.semanticscholar.org/CorpusID:278055935}
}

@ARTICLE{Dutta2023Vector,
  author = {Dutta, Neelotpal and Zhang, Tianyu and Fang, Guoxin and Yigit, Ismail E. and Wang, Charlie C. L.},
  title = {Vector Field-Based Volume Peeling for Multi-Axis Machining},
  journal = {J. Comput. Inf. Sci. Eng.},
  year = {2023},
  volume = {24},
  number = {5},
  pages = {051001},
  doi = {10.1115/1.4063861}
}

@ARTICLE{Ge2024Spiral,
  author = {Ge, Zhenghui and Hu, Qifan and Wang, Ziyi and Zhu, Yongwei},
  title = {Spiral tool path generation method for complex pocket machining based on electrostatic field theory},
  journal = {J. Braz. Soc. Mech. Sci. Eng.},
  year = {2024},
  volume = {46},
  number = {10},
  pages = {614},
  doi = {10.1007/s40430-024-05191-4}
}

@inproceedings{Goes2016Vector,
  author = {Goes, Fernando de and Desbrun, Mathieu and Tong, Yiying},
  title = {Vector field processing on triangle meshes},
  booktitle = {Proceedings of the ACM SIGGRAPH 2016 Courses},
  year = {2016},
  publisher = {Association for Computing Machinery},
  address = {Anaheim, California},
  pages = {Article 27, 49 pages},
  doi = {10.1145/2897826.2927303}
}

@ARTICLE{Grossmann2024Can,
  author = {Grossmann, Tamara G. and Komorowska, Urszula Julia and Latz, Jonas and Schönlieb, Carola-Bibiane},
  title = {Can physics-informed neural networks beat the finite element method?},
  journal = {IMA J. Appl. Math.},
  year = {2024},
  volume = {89},
  number = {1},
  pages = {143-174},
  doi = {10.1093/imamat/hxae011}
}

@ARTICLE{Hajdu2025Stable,
  author = {Hajdu, David and Franco, Oier and Sanz-Calle, Markel and Totis, Giovanni and Munoa, Jokin and Stepan, Gabor and Dombovari, Zoltan},
  title = {Stable tongues induced by milling tool runout},
  journal = {International Journal of Machine Tools and Manufacture},
  year = {2025},
  volume = {206},
  pages = {104258},
  doi = {10.1016/j.ijmachtools.2025.104258}
}

@ARTICLE{Hao2025Neural,
  author = {Hao, Jiancheng and Chen, Haokun and Hu, Pengcheng and He, Dong and Li, Yamin and Deng, Xiaoke and Lau, Tak Yu and Shi, Fan and Lu, Yanglong},
  title = {Neural surface partitioner: A physics-informed neural network for five-axis machining by a non-spherical cutting tool},
  journal = {J. Manuf. Syst.},
  year = {2025},
  volume = {83},
  pages = {976-991},
  doi = {10.1016/j.jmsy.2025.11.015}
}

@ARTICLE{Held2018On,
  author = {Held, Martin and de Lorenzo, Stefan},
  title = {On the generation of spiral-like paths within planar shapes},
  journal = {J. Comput. Des. Eng.},
  year = {2018},
  volume = {5},
  number = {3},
  pages = {348-357},
  doi = {10.1016/j.jcde.2017.11.011}
}

@ARTICLE{Kim2002Machining,
  author = {Kim, Bo H. and Choi, Byoung K.},
  title = {Machining efficiency comparison direction-parallel tool path with contour-parallel tool path},
  journal = {Comput.-Aided Des.},
  year = {2002},
  volume = {34},
  number = {2},
  pages = {89-95},
  doi = {10.1016/S0010-4485(00)00139-1}
}

@ARTICLE{Lee2003Contour,
  author = {Lee, Eungki},
  title = {Contour offset approach to spiral toolpath generation with constant scallop height},
  journal = {Comput.-Aided Des.},
  year = {2003},
  volume = {35},
  number = {6},
  pages = {511-518},
  doi = {10.1016/S0010-4485(01)00185-3}
}

@ARTICLE{Li2006Representing,
  author = {Li, Wan-chiu and Vallet, Bruno and Ray, Nicolas and Levy, Bruno},
  title = {Representing Higher-Order Singularities in Vector Fields on Piecewise Linear Surfaces},
  journal = {IEEE Trans. Vis. Comput. Graph.},
  year = {2006},
  volume = {12},
  number = {5},
  pages = {1315-1322},
  doi = {10.1109/TVCG.2006.173}
}

@ARTICLE{Li2025A,
  author = {Li, Dongyi and Lu, Yibin},
  title = {A method of numerical conformal mapping of bounded regions with a rectilinear slit},
  journal = {AIMS Math.},
  year = {2025},
  volume = {10},
  number = {10},
  pages = {8422-8445},
  doi = {10.3934/math.2025388}
}

@inproceedings{Lin2017Robot,
  author = {Lin, Yu-Yao and Ni, Chien-Chun and Lei, Na and Gu, Xianfeng David and Gao, Jie},
  title = {Robot Coverage Path planning for general surfaces using quadratic differentials},
  booktitle = {2017 IEEE International Conference on Robotics and Automation (ICRA)},
  year = {2017},
  publisher = {IEEE},
  address = {Singapore},
  pages = {5005-5011},
  doi = {10.1109/ICRA.2017.7989583}
}

@ARTICLE{Makhanov2022Vector,
  author = {Makhanov, Stanislav S.},
  title = {Vector fields for five-axis machining. A survey},
  journal = {Int. J. Adv. Manuf. Technol.},
  year = {2022},
  volume = {122},
  pages = {533-575},
  doi = {10.1007/s00170-022-09445-0}
}

@ARTICLE{Mali2021A,
  author = {Mali, Rahul A. and Gupta, T. V. K. and Ramkumar, J.},
  title = {A comprehensive review of free-form surface milling– Advances over a decade},
  journal = {J. Manuf. Process.},
  year = {2021},
  volume = {62},
  pages = {132-167},
  doi = {10.1016/j.jmapro.2020.12.014}
}

@ARTICLE{Marin2025Dimensional,
  author = {Marin, Felipe and González, Guillermo and de Lacalle, Luis Norberto López and Ortega, Naiara and Gómez-Escudero, Gaizka and Fernández-Lucio, Pablo and Del Olmo, Ander and González, Haizea},
  title = {Dimensional errors of aeronautical casings caused by machining of thin walls and features},
  journal = {Results Eng.},
  year = {2025},
  volume = {26},
  pages = {104719},
  doi = {10.1016/j.rineng.2025.104719}
}

@inproceedings{Mitra2024Singular,
  author = {Mitra, Rahul and Berumen, Erick Jimenez and Hofmann, Megan and Chien, Edward},
  title = {Singular Foliations for Knit Graph Design},
  booktitle = {Proceedings of the ACM SIGGRAPH 2024 Conference Papers},
  year = {2024},
  publisher = {Association for Computing Machinery},
  address = {Denver, CO, USA},
  pages = {Article 38, 11 pages},
  doi = {10.1145/3641519.3657487}
}

@ARTICLE{Nasser2011Numerical,
  author = {Nasser, Mohamed M. S.},
  title = {Numerical conformal mapping of multiply connected regions onto the second, third and fourth categories of Koebe's canonical slit domains},
  journal = {J. Math. Anal. Appl.},
  year = {2011},
  volume = {382},
  number = {1},
  pages = {47-56},
  doi = {10.1016/j.jmaa.2011.04.030}
}

@inproceedings{Rahaman2019On,
  author = {Rahaman, Nasim and Baratin, Aristide and Arpit, Devansh and Draxler, Felix and Lin, Min and Hamprecht, Fred and Bengio, Yoshua and Courville, Aaron},
  title = {On the Spectral Bias of Neural Networks},
  booktitle = {Proceedings of the 36th International Conference on Machine Learning},
  year = {2019},
  publisher = {PMLR},
  pages = {5301-5310}
}

@inproceedings{Rusinkiewicz2004Estimating,
  author = {Rusinkiewicz, S.},
  title = {Estimating curvatures and their derivatives on triangle meshes},
  booktitle = {Proceedings. 2nd International Symposium on 3D Data Processing, Visualization and Transmission},
  year = {2004},
  publisher = {IEEE},
  address = {Thessaloniki, Greece},
  pages = {486-493},
  doi = {10.1109/TDPVT.2004.1335277}
}

@inproceedings{Shen2019Networks,
  author = {Shen, Zhengyang and Han, Xu and Xu, Zhenlin and Niethammer, Marc},
  title = {Networks for Joint Affine and Non-parametric Image Registration},
  booktitle = {Proceedings of the IEEE/CVF Conference on Computer Vision and Pattern Recognition (CVPR)},
  year = {2019},
  publisher = {IEEE},
  pages = {4224-4233}
}

@ARTICLE{Shen2024Spiral,
  author = {Shen, Changqing and Mao, Sihao and Xu, Bingzhou and Wang, Ziwei and Zhang, Xiaojian and Yan, Sijie and Ding, Han},
  title = {Spiral complete coverage path planning based on conformal slit mapping in multi-connected domains},
  journal = {Int. J. Robot. Res.},
  year = {2024},
  volume = {43},
  number = {14},
  pages = {2183-2203},
  doi = {10.1177/02783649241251385}
}

@techreport{Shen2025Conformal,
  author = {Shen, Changqing and Xu, BingZhou and Zhang, Xiaojian and Yan, Sijie and Ding, Han},
  title = {Conformal Slit Mapping Based Spiral Tool Trajectory Planning for Ball-end Milling on Complex Freeform Surfaces},
  year = {2025},
  url = {https://doi.org/10.48550/arXiv.2504.06310},
  note = {arXiv preprint arXiv:2504.06310}
}

@ARTICLE{Song2018Epsilon,
  author = {Song, Junnan and Gupta, Shalabh},
  title = {$\varepsilon^\star$: An Online Coverage Path Planning Algorithm},
  journal = {IEEE Trans. Robot.},
  year = {2018},
  volume = {34},
  number = {2},
  pages = {526-533},
  doi = {10.1109/TRO.2017.2780259}
}

@ARTICLE{Sun2016Smooth,
  author = {Sun, Yuwen and Xu, Jinting and Jin, Chunning and Guo, Dongming},
  title = {Smooth tool path generation for 5-axis machining of triangular mesh surface with nonzero genus},
  journal = {Comput.-Aided Des.},
  year = {2016},
  volume = {79},
  pages = {60-74},
  doi = {10.1016/j.cad.2016.06.001}
}

@ARTICLE{Vekhter2019Weaving,
  author = {Vekhter, Josh and Zhuo, Jiacheng and Gil Fandino, Luisa F and Huang, Qixing and Vouga, Etienne},
  title = {Weaving geodesic foliations},
  journal = {ACM Trans. Graph.},
  year = {2019},
  volume = {38},
  number = {4},
  pages = {Article 34, 22 pages},
  doi = {10.1145/3306346.3323043}
}

@ARTICLE{Wang2025Tool,
  author = {Wang, Shu and Zhang, Hui and Li, Bingran and Ye, Peiqing},
  title = {Tool trajectory planning with inter-path consistency of feedrate for freeform surface machining},
  journal = {CIRP J. Manuf. Sci. Technol.},
  year = {2025},
  volume = {62},
  pages = {1-15},
  doi = {10.1016/j.cirpj.2025.08.003}
}

@ARTICLE{Wu2019Energy,
  author = {Wu, Chenming and Dai, Chengkai and Gong, Xiaoxi and Liu, Yong-Jin and Wang, Jun and Gu, Xianfeng David and Wang, Charlie C. L.},
  title = {Energy-Efficient Coverage Path Planning for General Terrain Surfaces},
  journal = {IEEE Robot. Autom. Lett.},
  year = {2019},
  volume = {4},
  number = {3},
  pages = {2584-2591},
  doi = {10.1109/LRA.2019.2899920}
}

@ARTICLE{Wu2023Numerical,
  author = {Wu, Kang and Lu, Yibin},
  title = {Numerical computation of preimage domains for spiral slit regions and simulation of flow around bodies},
  journal = {Math. Biosci. Eng.},
  year = {2023},
  volume = {20},
  number = {1},
  pages = {720-736},
  doi = {10.3934/mbe.2023033}
}

@ARTICLE{Wu2024Pose,
  author = {Wu, Lei and Zang, Xizhe and Yin, Wenxin and Zhang, Xuehe and Li, Changle and Zhu, Yanhe and Zhao, Jie},
  title = {Pose and Path Planning for Industrial Robot Surface Machining Based on Direction Fields},
  journal = {IEEE Robot. Autom. Lett.},
  year = {2024},
  volume = {9},
  number = {11},
  pages = {10455-10462},
  doi = {10.1109/LRA.2024.3474521}
}

@ARTICLE{Xu2019Contour,
  author = {Xu, Chen-Yang and Li, Jing-Rong and Wang, Qing-Hui and Hu, Guang-Hua},
  title = {Contour parallel tool path planning based on conformal parameterisation utilising mapping stretch factors},
  journal = {Int. J. Prod. Res.},
  year = {2019},
  volume = {57},
  number = {1},
  pages = {1-15},
  doi = {10.1080/00207543.2018.1456699}
}

@inproceedings{Xu2022Global,
  author = {Xu, Xiaohu and Ye, Songtao and Yang, Zeyuan and Yan, Sijie and Ding, Han},
  title = {Global optimal trajectory planning of mobile robot grinding for high-speed railway body},
  booktitle = {Intelligent Robotics and Applications. ICIRA 2022},
  year = {2022},
  editor = {Liu, Honghai and Yin, Zhouping and Liu, Lianqing and Jiang, Li and Gu, Guoying and Wu, Xinyu and Ren, Weihong},
  publisher = {Springer, Cham},
  doi = {10.1007/978-3-031-13835-5_44}
}

@ARTICLE{Yan2019An,
  author = {Yan, Sijie and Xu, Xiaohu and Yang, Zeyuan and Zhu, Dahu and Ding, Han},
  title = {An improved robotic abrasive belt grinding force model considering the effects of cut-in and cut-off},
  journal = {J. Manuf. Process.},
  year = {2019},
  volume = {37},
  pages = {496-508},
  doi = {10.1016/j.jmapro.2018.12.029}
}

@ARTICLE{Yang2023Template,
  author = {Yang, Tong and Miro, Jaime Valls and Nguyen, Minh and Wang, Yue and Xiong, Rong},
  title = {Template-Free Nonrevisiting Uniform Coverage Path Planning on Curved Surfaces},
  journal = {IEEE/ASME Trans. Mechatron.},
  year = {2023},
  volume = {28},
  number = {4},
  pages = {1853-1861},
  doi = {10.1109/TMECH.2023.3275214}
}

@ARTICLE{Yang2024Dynamic,
  author = {Yang, Zeyuan and Xu, Xiaohu and Kuang, Minxing and Zhu, Dahu and Yan, Sijie and Ge, Shuzhi Sam and Ding, Han},
  title = {Dynamic compliant force control strategy for suppressing vibrations and over-grinding of robotic belt grinding system},
  journal = {IEEE Transactions on Automation Science and Engineering},
  year = {2024},
  volume = {21},
  number = {3},
  pages = {4536-4547},
  doi = {10.1109/TASE.2023.3298357}
}

@ARTICLE{Zhang2006Vector,
  author = {Zhang, Eugene and Mischaikow, Konstantin and Turk, Greg},
  title = {Vector field design on surfaces},
  journal = {ACM Trans. Graph.},
  year = {2006},
  volume = {25},
  number = {4},
  pages = {1294-1326},
  doi = {10.1145/1183287.1183290}
}

@ARTICLE{Zhao2024Hybrid,
  author = {Zhao, Tao and Yan, Zhaoyang and Wang, Liwei and Pan, Rui and Wang, Xiaowei and Liu, Kun and Guo, Kaiwei and Hu, Qingsong and Chen, Shujun},
  title = {Hybrid path planning method based on skeleton contour partitioning for robotic additive manufacturing},
  journal = {Robot. Comput.-Integr. Manuf.},
  year = {2024},
  volume = {85},
  pages = {102633},
  doi = {10.1016/j.rcim.2023.102633}
}

@ARTICLE{Zhuang2010High,
  author = {Zhuang, Chungang and Xiong, Zhenhua and Ding, Han},
  title = {High speed machining tool path generation for pockets using level sets},
  journal = {Int. J. Prod. Res.},
  year = {2010},
  volume = {48},
  number = {19},
  pages = {5749-5766},
  doi = {10.1080/00207540903232771}
}

@ARTICLE{Zou2021Length,
  author = {Zou, Qiang},
  title = {Length-optimal tool path planning for freeform surfaces with preferred feed directions based on Poisson formulation},
  journal = {Comput.-Aided Des.},
  year = {2021},
  volume = {139},
  pages = {103072},
  doi = {10.1016/j.cad.2021.103072}
}

@ARTICLE{Zou2014Iso,
  author = {Zou, Qiang and Zhang, Juyong and Deng, Bailin and Zhao, Jibin},
  title = {Iso-level tool path planning for free-form surfaces},
  journal = {Comput.-Aided Des.},
  year = {2014},
  volume = {53},
  pages = {117-125},
  doi = {10.1016/j.cad.2014.04.006}
}

@InProceedings{Itakura2020,
author="Savchenko, Andrey V.
and Savchenko, Vladimir V.
and Savchenko, Lyudmila V.",
editor="Kononov, Alexander
and Khachay, Michael
and Kalyagin, Valery A
and Pardalos, Panos",
title="Optimization of Gain in Symmetrized Itakura-Saito Discrimination for Pronunciation Learning",
booktitle="Mathematical Optimization Theory and Operations Research",
year="2020",
publisher="Springer International Publishing",
address="Cham",
pages="440--454",
isbn="978-3-030-49988-4"
}

@ARTICLE{abrahamsen2019cuthole,
title = {Spiral tool paths for high-speed machining of 2D pockets with or without islands},
journal = {Journal of Computational Design and Engineering},
volume = {6},
number = {1},
pages = {105-117},
year = {2019},
issn = {2288-4300},
doi = {https://doi.org/10.1016/j.jcde.2018.01.003},
url = {https://www.sciencedirect.com/science/article/pii/S2288430017302452},
author = {Mikkel Abrahamsen},
}

@ARTICLE{patel2017decomposition,

    title = {Quantitative Comparison of Pocket Geometry and Pocket Decomposition to Obtain Improved Spiral Tool Path: A Novel Approach},
    journal = {Journal of Manufacturing Science and Engineering},
    volume = {139},
    number = {3},
    pages = {031020},
    year = {2017},
    month = {01},
    issn = {1087-1357},
    doi = {10.1115/1.4034896},
    url = {https://doi.org/10.1115/1.4034896},
    author = {Patel, Divyangkumar D. and Lalwani, Devdas I.},
}



\printcredits

\appendix
\renewcommand{\thefigure}{A\arabic{figure}}
\renewcommand{\theequation}{A\arabic{equation}}
\setcounter{figure}{0}
\setcounter{equation}{0}

\section{The Relation Between Scallop Height and the Scalar-Field Gradient}

As shown in Figures~\ref{fig:relationships}(a) and (b), let $P_i$ and $P_{i+1}$ denote two adjacent cutter contact points on the surface $S$. They lie on two adjacent isoparametric lines of the scalar field $T$, denoted as $CL_i=\{P\in S|T(P)=T_i\}$  and $CL_{i+1}=\{P\in S|T(P)=T_{i+1}\}$, respectively. The corresponding ball-end mill centers were offset from the surface along the normal direction by $R_c$, where $R_c = \frac{1}{K_c}$ is the radius of the ball-end mill. The machining scallop height between $P_i$ and $P_{i+1}$ can be expressed as follows \cite{Zou2014Iso}:
\begin{equation}
h = \frac{K_s + K_c}{8} \left\| \overrightarrow{P_{i+1} P_i} \right\|^2 + o\left( \left\| \overrightarrow{P_{i+1} P_i} \right\|^3 \right)
\label{eq:h_relate_dist}
\end{equation}

where $K_s$ denotes the normal curvature of the surface $S$ at $P_i$ in the direction of $\overrightarrow{P_{i} P_{i+1}}$, and $\overrightarrow{P_{i} P_{i+1}}$ is the vector between $P_i$ and $P_{i+1}$ (Figure~\ref{fig:relationships}(b)). The notation $o\left( \left\| \overrightarrow{P_{i+1} P_i} \right\|^3 \right)$ represents the cubic infinitesimal of the Euclidean norm of $\overrightarrow{P_{i} P_{i+1}}$.

When $P_{i+1}$ is sufficiently close to $P_i$, the direction of $\overrightarrow{P_{i} P_{i+1}}$ aligns with the gradient of $T$ at $P_i$, which can be denoted as $\nabla T(P_i)$. According to Taylor’s theorem, the following equation can be derived:
\begin{equation}
\left( \left\| \nabla T(P_i) \right\| \right) \left( \left\| \overrightarrow{P_i P_{i+1}} \right\| \right) + o\left( \left\| \overrightarrow{P_i P_{i+1}} \right\|_2^2 \right) = |T_{i+1} - T_i|
\label{eq:dt_relate_dist}
\end{equation}

By neglecting higher-order infinitesimals and combining Equations ~\ref{eq:h_relate_dist} and ~\ref{eq:dt_relate_dist}, the following equation can be obtained:
\begin{equation}
h = \left| T_{i+1} - T_i \right|^2 \frac{K_s + K_c}{8\left\| \nabla T(P_i) \right\|_2^2}
\label{eq:h_relate_dt}
\end{equation}

Because $| T_{i+1}-T_i|^2$ is constant, it follows from Equation~\ref{eq:h_relate_dt} that the variation in the scallop height between adjacent isocurves can be characterized by the changes in $\frac{K_s+K_c}{8\|\nabla T \|^2_2}$. The global fluctuations of this quantity about $Q_S$ can be evaluated by Equation~\ref{eq:Ew}.

\section{The Discrete Form of $E_w$}

The discrete form of Equation~\ref{eq:Ew} on a triangular mesh is as follows:
\begin{equation}
    E_{w}=\sum_{i \in|F|} \boldsymbol{A}_{i}\left(\frac{K_{s}\left(A_{i}\right)+K_{c}}{8Q_S\left\|\nabla T\left(A_{i}\right)\right\|_{2}^{2}}+\frac{8Q_S\left\|\nabla T\left(A_{i}\right)\right\|_{2}^{2}}{K_{s}\left(A_{i}\right)+K_{c}}\right)
\end{equation}

where
\begin{equation}
  Q_S= \frac{\sum_{i \in|F|} \frac{A_i(K_s(A_i)+K_c)}{8\,\|\nabla T_i(A_i)\|_2^2}}
  {\sum_{i \in|F|}A_i}
\end{equation}

where $F$ denotes the set of triangular mesh faces, and $i$ represents the index of the set elements. For $A_i$, denotes the area of the $i$-th mesh face; and $\nabla T(A_i)$ and $K_s(A_i)$ represent the scalar-field gradient and the discrete curvature along the gradient direction on the corresponding triangular face, respectively. The discrete curvature $K_s(A_i)$ can be determined as:
\begin{equation}
K_{s}\left(\boldsymbol{A}_{i}\right)=\left(\frac{\nabla T\left(A_{i}\right)^{\mathrm{T}}}{\left\|\nabla T\left(A_{i}\right)\right\|}\right) \boldsymbol{C}_{T}\left(\frac{\nabla T\left(A_{i}\right)}{\left\|\nabla T\left(A_{i}\right)\right\|}\right)
\end{equation}

where $C_T$ denotes the curvature tensor on the mesh face, which can be computed following reference \cite{Rusinkiewicz2004Estimating}.

\section{The Discrete Form of $E_k$}
Let $K_f$ be the curvature of the iso-level curve $CL$, which can be decomposed into two components along the unit orthogonal vectors. The normal curvature vector $\overrightarrow{n}_{n}$ and the geodesic curvature vector $\overrightarrow{n}_{g}$ are denoted as $K_n$ and $K_g$, respectively. The relationship between these components is expressed as follows:
\begin{equation}
    K_{f}^{2} \overrightarrow{{n}}_{f}=K_{n}^{2} \overrightarrow{{n}}_{n}+K_{g}^{2} \overrightarrow{{n}}_{g}
\end{equation}

where $\overrightarrow{n}_{n}$ and $\overrightarrow{n}_{g}$ are perpendicular to and lie within the tangent plane of $S$, respectively. The relationships among $K_f$, $K_g$, and $K_n$ are illustrated in Figures~\ref{fig:relationships}(a), (c), and (d), respectively.

Based on this, Equation~\ref{eq:Ek} can be expressed as:
\begin{equation}
    E_{k}=E_{n}+E_{g}=\int_{S} K_{n}^{2} d S+\int_{S} K_{g}^{2} d S
\end{equation}

The discrete form of $E_n$ can be expressed as:
\begin{equation}
    E_{n}=\sum_{i \in|F|} {A}_{i}\left(\frac{\nabla T\left(A_{i}\right)}{\left\|\nabla T\left(A_{i}\right)\right\|}\right)^{\mathrm{T}} {C}_{T}^{\mathrm{T}}\left(\frac{\nabla T\left(A_{i}\right)}{\left\|\nabla T\left(A_{i}\right)\right\|}\right)
\end{equation}

\begin{figure}
  \centering
  \includegraphics[width=0.8\linewidth]{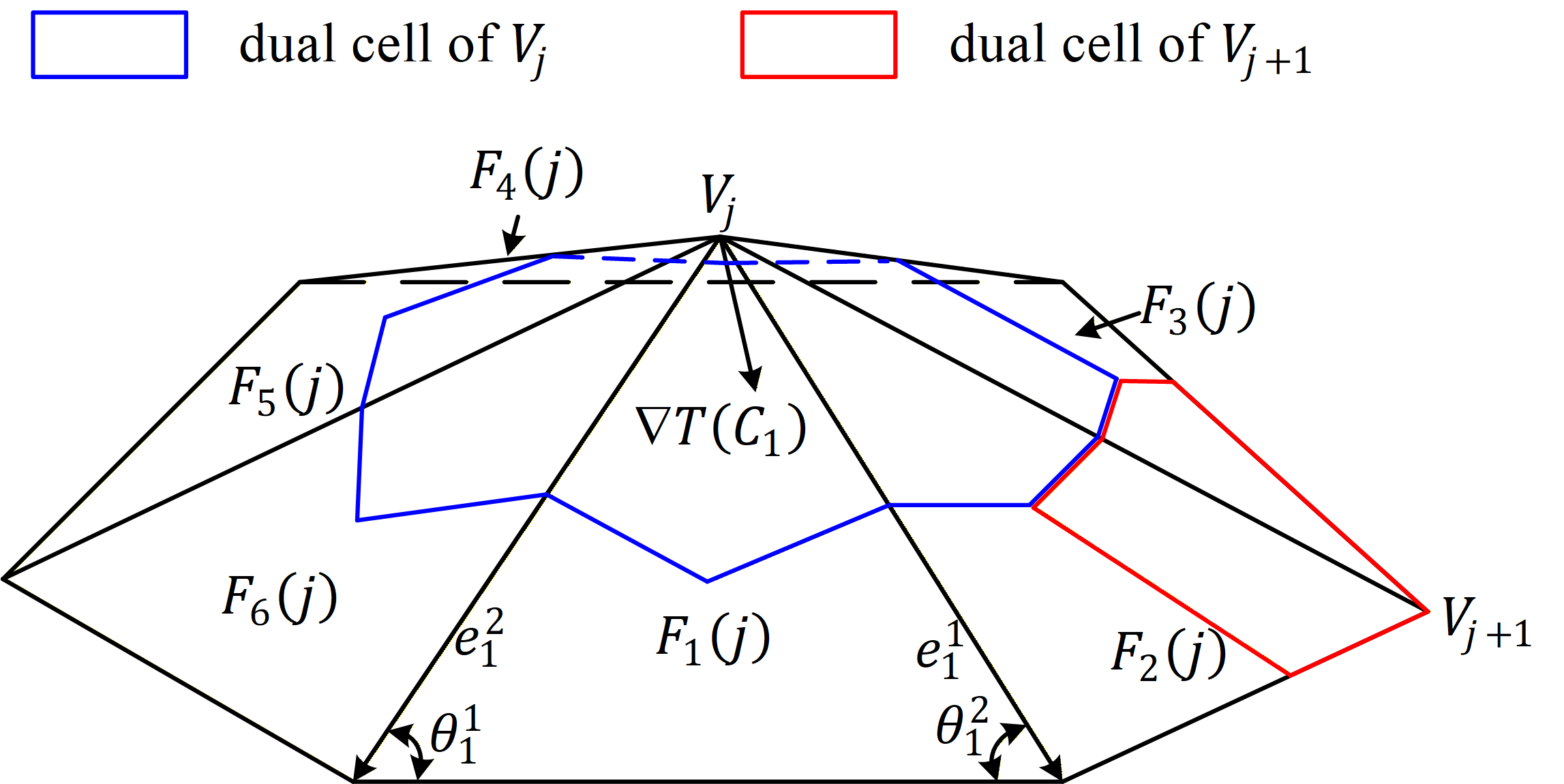}
  \caption{Dual cells of the vertices and the relationships among $\nabla T$, $\theta_{k}^{1}$, $\theta_{k}^{2}$, $e_k^1$, and $e_k^2$ on the triangular faces.}
  \label{fig:dualcells}
\end{figure}

The discrete form of $E_g$ can be expressed as:
\begin{equation}
    E_{g}=\sum_{j \in|V|} C_{j} div\left(\frac{\nabla T\left(C_{j}\right)}{\left\|\nabla T\left(C_{j}\right)\right\|}\right)^2
    \label{eg_discrete}
\end{equation}

where $V$ denotes the set of mesh vertices; $C_j$ represents the area of the dual cell corresponding to $j$--th vertex; and $div\left(\frac{\nabla T\left(C_{j}\right)}{\left\|\nabla T\left(C_{j}\right)\right\|}\right)$ is the divergence of the scalar field at the vertex, which can be computed as follows:
\begin{equation}
\begin{aligned}
\operatorname{div}\left(
\frac{\nabla T(C_j)}{\lVert \nabla T(C_j) \rVert}
\right)
&= \frac{1}{2C_j} \sum_{k \in F(j)}
\frac{
\cot\theta_k^1 \bigl( \boldsymbol{e}_k^1 \cdot \nabla T(C_k) \bigr)
}{\lVert \nabla T(C_k) \rVert}
\\
&\quad +
\frac{
\cot\theta_k^2 \bigl( \boldsymbol{e}_k^2 \cdot \nabla T(C_k) \bigr)
}{\lVert \nabla T(C_k) \rVert}
\end{aligned}
\end{equation}

where $F(j)$ denotes the set of all triangular faces containing vertex $j$. The relationships among $\theta_{k}^{1}$, $\theta_{k}^{2}$, $e_k^1$, and $e_k^2$ on the $k$-th triangular face are illustrated in Figure~\ref{fig:dualcells}.

\section{Details of Computing $T_{\mathrm{init}}$}

Given $\Theta$, the corresponding conformal slit mapping is determined. The scalar field $T$ on $S$, together with its associated energy $E$, is then fully determined by the unknown function $f$. The objective is therefore to find the optimal function $f_{\min}$ that minimizes $E$, which is formulated as
\begin{equation}
\begin{cases}
E_{\min}(\Theta) := \min_{f \in \{f\}} E\bigl(T(\Theta,f)\bigr), \\
f_{\min}(\Theta) := \arg\min_{f \in \{f\}} E\bigl(T(\Theta,f)\bigr).
\end{cases}
\end{equation}
Here, $\{f\}$ denote the function space satisfying the prescribed constraints in Equation~\ref{eq:fconstraint}.

The optimal function $f_{\min}$ and the corresponding minimum energy $E_{\min}$ are obtained numerically using a perturbation-based method. In this method, $f$ is represented by linear interpolation over a set of discrete nodes, with the first node $f\bigl(\min(\|S^S\|)\bigr)$ fixed. At each iteration, a perturbation $\delta f$ is introduced to the remaining nodes so as to reduce the energy $E$ while preserving the monotonicity constraint $f'>0$, until $f$ converges to $f_{\min}$, as illustrated in Figure~\ref{fig:fconverge}. The detailed implementation is given in Appendix C of \cite{Shen2025Conformal} and is therefore omitted here.
\begin{figure}
  \centering
  \includegraphics[width=0.8\linewidth]{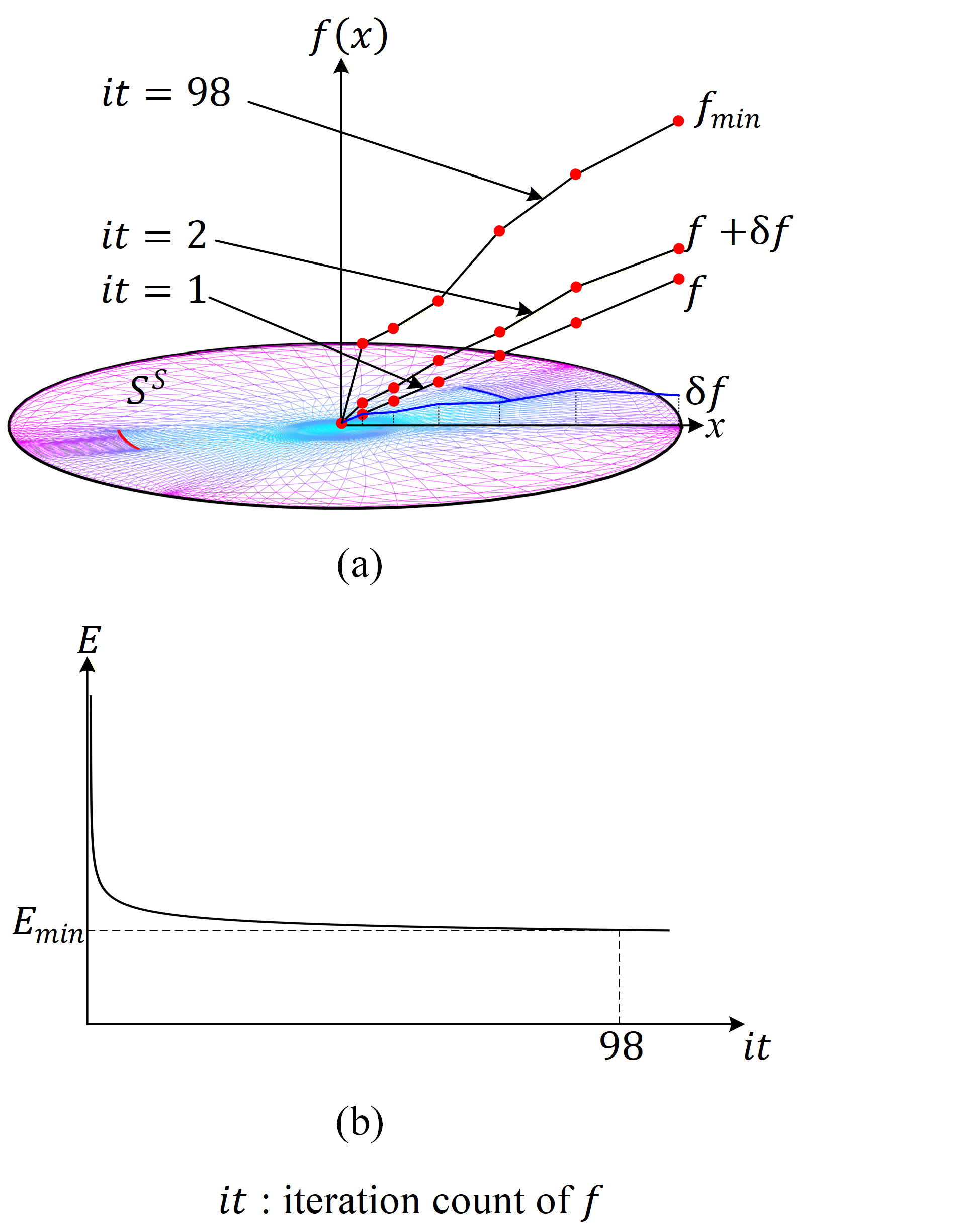}
  \caption{\label{fig:fconverge}(a) Linearized discretization and iterative refinement of the unknown function $f$. (b) Variation of the energy $E$ during the optimization of $f$.}
\end{figure}

Thus, for a given $\Theta$, the corresponding optimal function $f_{\min}$ and the minimum energy $E_{\min}(\Theta)$ can be obtained automatically. The next step is to determine the optimal $\Theta$ that minimizes $E_{\min}$:
\begin{equation}
\begin{cases}
E_{\min}^{\mathrm{opt}} := \min_{\Theta \in \{S\setminus\Gamma\} \cup \{\Gamma_i \mid i=1,2,\dots,m\}} E_{\min}(\Theta), \\
\Theta^{\mathrm{opt}} := \arg\min_{\Theta \in \{S\setminus\Gamma\} \cup \{\Gamma_i \mid i=1,2,\dots,m\}} E_{\min}(\Theta).
\end{cases}
\end{equation}

\begin{figure}
  \centering
  \includegraphics[width=0.8\linewidth]{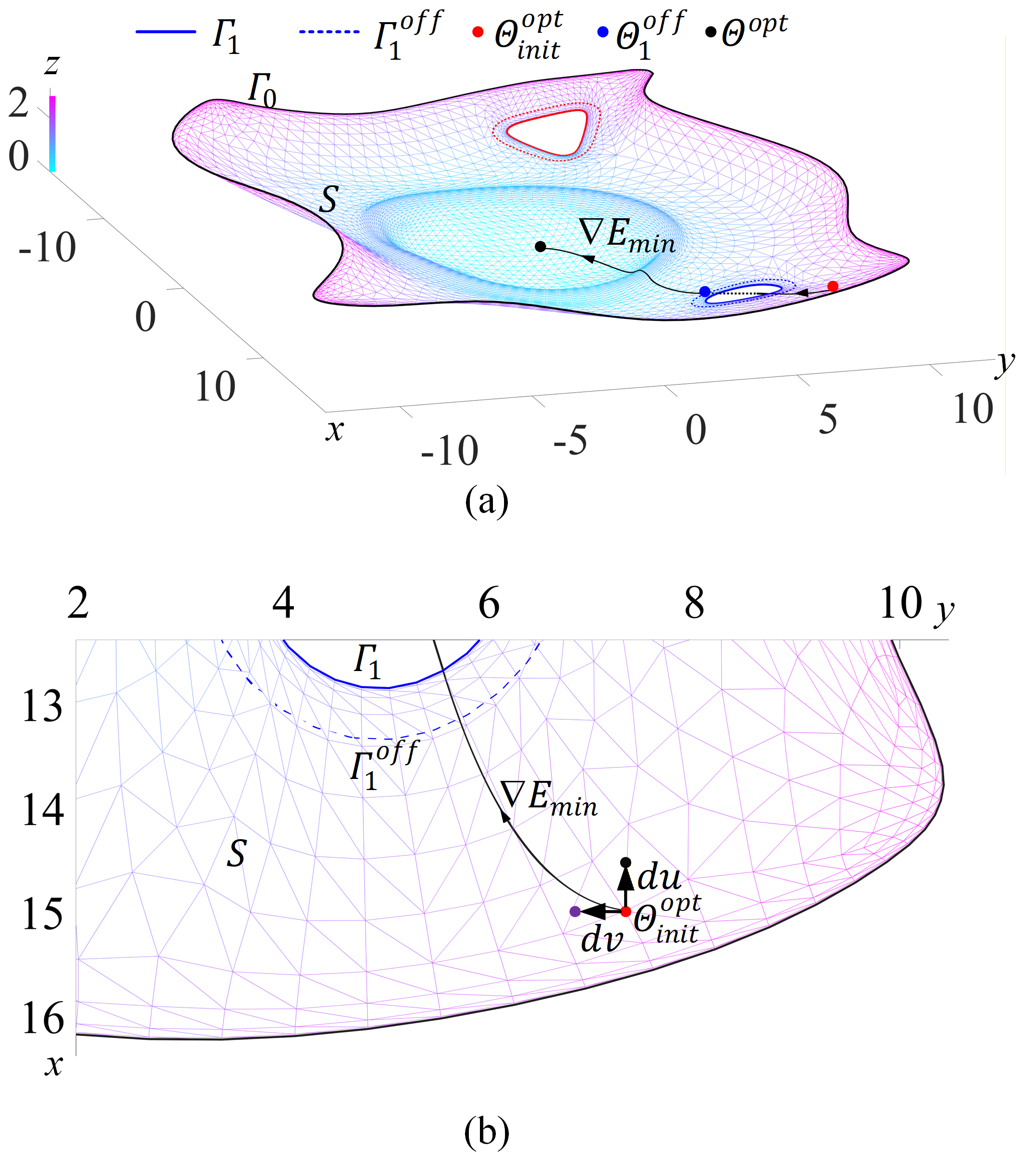}
  \caption{\label{fig:thetagradient}Gradient-descent search for $\Theta^{\mathrm{opt}}$. (a) overview. (b) local enlarged drawing.}
\end{figure}

As illustrated in Figure~\ref{fig:thetagradient}, the optimal element $\Theta^{\mathrm{opt}}$ is obtained through a gradient-descent search. Starting from an arbitrary initial point $\Theta_{\mathrm{init}}^{\mathrm{opt}}$ on $S\setminus\Gamma$, two perturbed points, $\Theta_{\mathrm{init}}^{\mathrm{opt}}+du$ and $\Theta_{\mathrm{init}}^{\mathrm{opt}}+dv$, are sampled along the two axes of the local coordinate system, as shown in Figure~\ref{fig:thetagradient}(b). Here, $du$ and $dv$ denote small positional perturbations along the local coordinate directions. The corresponding energy values $E_{\min}(\Theta_{\mathrm{init}}^{\mathrm{opt}}+du)$ and $E_{\min}(\Theta_{\mathrm{init}}^{\mathrm{opt}}+dv)$ are then evaluated to approximate the gradient $\nabla E_{\min}$, and $\Theta$ is iteratively updated along the negative gradient direction.

When $\Theta$ crosses $\Gamma_i$ and leaves $S$ during the search, the gradient $\nabla E_{\min}$ becomes undefined. In such cases, $\Theta$ is transferred to $\Theta_i^{\mathrm{off}}$ to allow the iteration to continue, where $\Theta_i^{\mathrm{off}}$ is defined by
\begin{equation}
\Theta_i^{\mathrm{off}} := \arg\min_{\Theta \in \Gamma_i^{\mathrm{off}}} E_{\min}(\Theta).
\end{equation}

Here, $\Gamma_i^{\mathrm{off}}$ denotes a curve obtained by offsetting $\Gamma_i$ slightly toward the interior of $S$. The iteration terminates in two cases. In the first case, $\Theta$ converges to an interior point, and $\Theta^{\mathrm{opt}}$ is taken as that point. In the second case, $\Theta$ repeatedly crosses an inner boundary $\Gamma_i$, in which case $\Theta^{\mathrm{opt}}$ is taken as the corresponding boundary component.

\begin{figure*}
  \centering
  \includegraphics[width=0.7\linewidth]{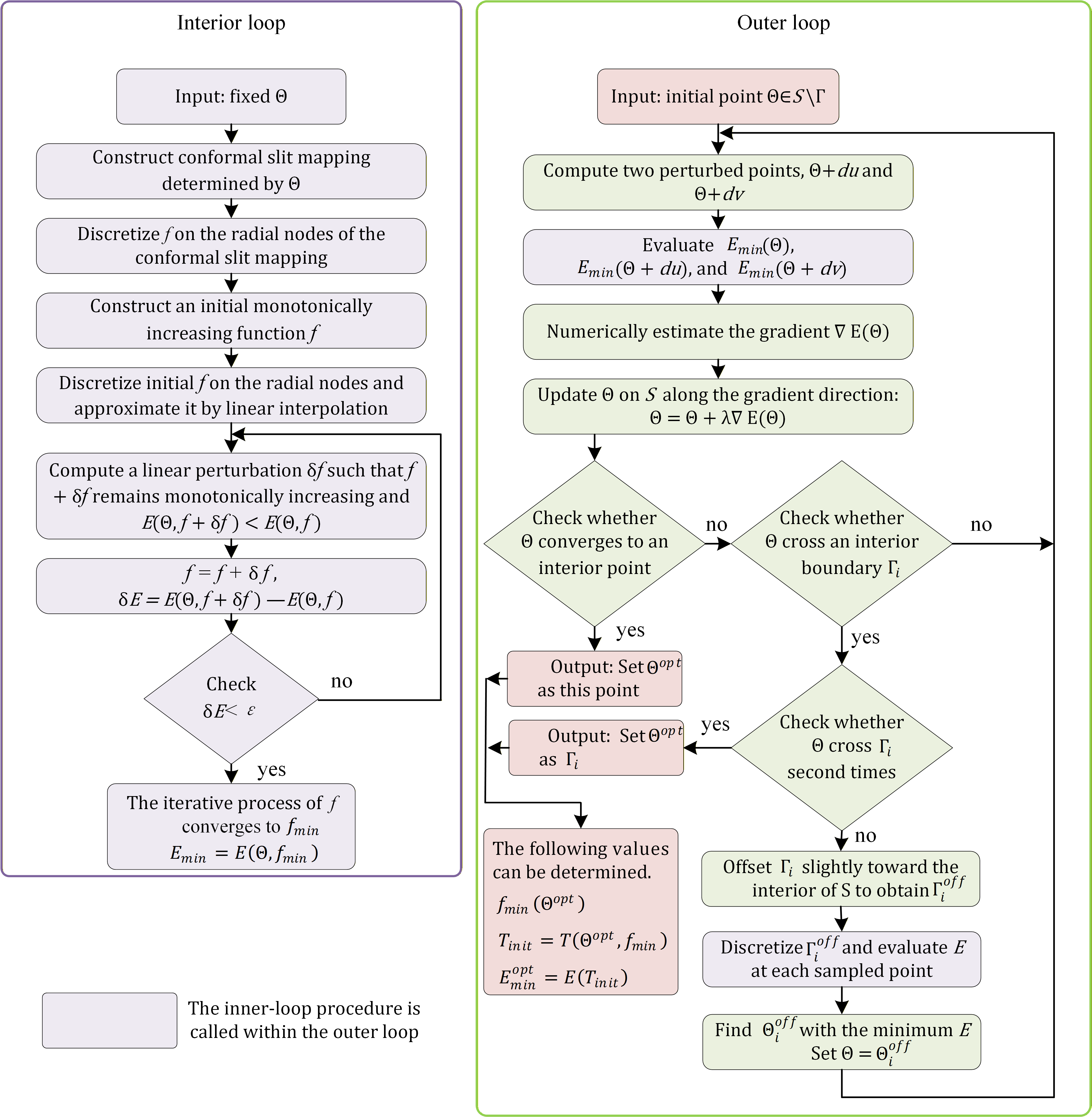}
  \caption{\label{fig:procedure}The procedure for computing $\Theta^{\mathrm{opt}}$, $T_{init}$ and $E_{\min}^{\mathrm{opt}}$}
\end{figure*}

In this manner, for a given surface $S$, $\Theta^{\mathrm{opt}}$ and the corresponding optimal function $f_{\min}$ can be determined such that
\begin{equation}
E_{\min}^{\mathrm{opt}} = E\bigl(T(\Theta^{\mathrm{opt}},f_{\min})\bigr).
\end{equation}
Consequently, $T\bigl(\Theta^{\mathrm{opt}},f_{\min}\bigr)$ serves as the desired initial scalar field $T_{\mathrm{init}}$. The procedure for computing $\Theta^{\mathrm{opt}}$, $T_{init}$ and $E_{\min}^{\mathrm{opt}}$ is illustrated in Figure~\ref{fig:procedure}.

\end{document}